\documentclass[journal]{IEEEtran}

\usepackage{multirow}
\usepackage{booktabs}
\usepackage{color,colortbl}
\usepackage{caption}
\usepackage{subcaption}
\usepackage{pifont}
\newcommand{\cmark}{\text{\ding{51}}}
\newcommand{\xmark}{\text{\ding{55}}}

\definecolor{Gray}{gray}{0.9}
\definecolor{Blue}{rgb}{0, 0.4470, 0.7410}
\definecolor{Red}{rgb}{0.8500, 0.3250, 0.0980}
\newcommand\minisection[1]{\noindent \textbf{#1}}
\newcommand\hl[1]{#1}
\usepackage[pagebackref=true,breaklinks=true,colorlinks,bookmarks=false]{hyperref}

\usepackage{cite}

\ifCLASSINFOpdf
   \usepackage[pdftex]{graphicx}
\else
\fi

\usepackage{adjustbox}
\usepackage{amsmath,amssymb}

\begin{document}
\title{Efficient Few-Shot Object Detection via Knowledge Inheritance}

\author{Ze~Yang,
        Chi~Zhang,
        Ruibo~Li,
        Yi~Xu,
        and~Guosheng~Lin
\thanks{Ze~Yang, Chi~Zhang, Ruibo~Li, and~Guosheng~Lin are with School of Computer Science and Engineering, Nanyang Technological University (NTU), Singapore 639798.
Yi~Xu is with OPPO US Research Center, InnoPeak Technology, Inc.
(e-mail: \{ze001, chi007, ruibo001\}@e.ntu.edu.sg, yi.xu@innopeaktech.com, gslin@ntu.edu.sg).
Corresponding author: Guosheng Lin}
}

\markboth{Journal of \LaTeX\ Class Files,~Vol.~14, No.~8, August~2015}%
{Shell \MakeLowercase{\textit{et al.}}: Bare Demo of IEEEtran.cls for IEEE Journals}

\maketitle

\begin{abstract}
Few-shot object detection (FSOD), which aims at learning a generic detector that can adapt to unseen tasks with scarce training samples, has witnessed consistent improvement recently. However, most existing methods ignore the \textbf{efficiency} issues, \textit{e.g.}, high computational complexity and slow adaptation speed.
Notably, efficiency has become an increasingly important evaluation metric for few-shot techniques due to an emerging trend toward embedded AI. 
To this end, we present an efficient pretrain-transfer framework (PTF) baseline with no computational increment, which achieves comparable results with previous state-of-the-art (SOTA) methods.
Upon this baseline, we devise an initializer named knowledge inheritance (KI) to reliably initialize the novel weights for the box classifier, which effectively facilitates the knowledge transfer process and boosts the adaptation speed.
Within the KI initializer, we propose an adaptive length re-scaling (ALR) strategy to alleviate the vector length inconsistency between the predicted novel weights and the pretrained base weights.
Finally, our approach not only achieves the SOTA results across three public benchmarks, \textit{i.e.}, PASCAL VOC, COCO and LVIS, but also exhibits high efficiency with 
$1.8\text{\textendash}100\times$ faster adaptation speed against the other methods on COCO/LVIS benchmark during few-shot transfer.
To our best knowledge, this is the \textbf{first work} to consider the efficiency problem in FSOD. We hope to motivate a trend toward powerful yet efficient few-shot technique development. The codes are publicly available at \url{https://github.com/Ze-Yang/Efficient-FSOD}.

\end{abstract}


\IEEEpeerreviewmaketitle

\section{Introduction}\label{sec:introduction}
\IEEEPARstart{O}{bject} detection has witnessed significant breakthrough~\cite{girshick2014rich,girshick2015fast,Ren:2017:fasterrcnn,liu2016ssd,tian2019fcos} with the recent advancement of deep neural network techniques. 
Nevertheless, the promising performance usually comes at the cost of massive human resources to obtain large-scale annotated datasets for network training. As a result, few-shot object detection (FSOD) was developed to alleviate this issue by learning a general detector that can be adapted to unseen tasks with only limited annotated samples.

Recently, FSOD has drawn increasing interest from the computer vision community with great efforts being placed on improving the performance. However, most of the existing methods~\cite{kang2019few,yan2019meta,fan2020few,xiao2020few,yang2020context} consistently ignore a crucial problem, \textit{i.e.}, \textit{efficiency}.
In the past, this was not a critical issue since few-shot learning was generally conducted on the powerful cloud servers. 
Nowadays, with the rapid development of electronic technologies,  mobile devices become much more computationally powerful, which makes it possible to perform inference and even lightweight training on such devices. As a result, the advancement incubates some real-world few-shot applications, \textit{e.g.}, online incremental learning~\cite{li2019rilod}, onsite model adaptation~\cite{song2018situ}, etc.
Nevertheless, given the moderate computational capacity of mobile devices, these applications indeed raise an \textit{efficiency} concern for the few-shot techniques.

\begin{figure}[t]
    \centering
    \includegraphics[width=\linewidth]{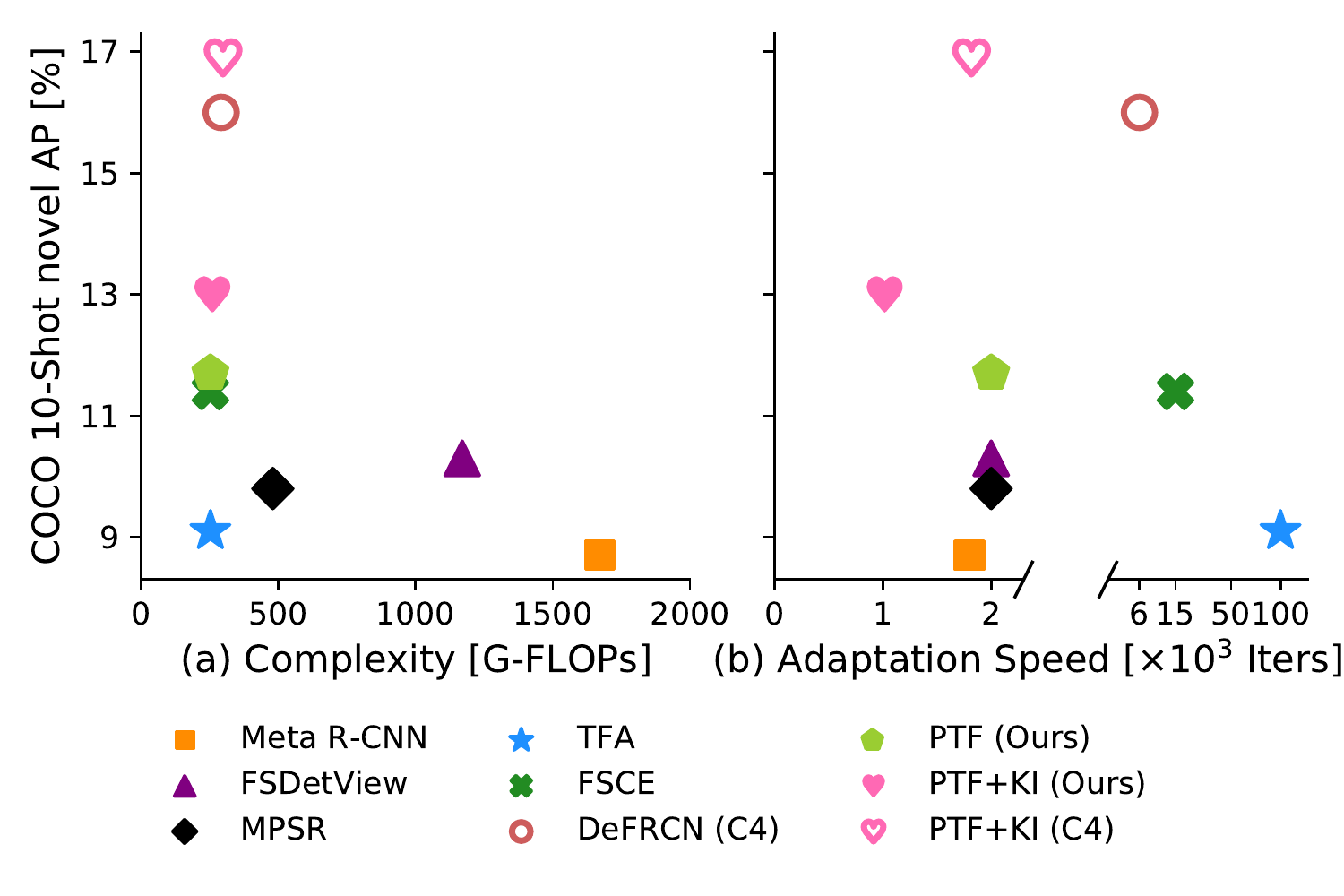}
    \caption{Comparison of performance vs. efficiency (computational complexity and adaptation speed) during few-shot transfer under COCO 10-shot setting. The proposed PTF+KI consistently outperforms the other methods on novel AP with much higher efficiency: (a) taking up only $15.1\%$, $21.6\%$ and $52.6\%$ FLOPs of those complicated methods with extra components or inputs, \textit{i.e.}, Meta R-CNN~\cite{yan2019meta}, FSDetView~\cite{xiao2020few} and MPSR~\cite{wu2020multi} respectively; (b) boosting the adaptation speed by $1.8\text{\textendash}100\times$ against the competing methods. The hollow shape indicates that the method uses ResNet101-C4 instead of ResNet101-FPN as the network architecture for the base detector Faster R-CNN~\cite{Ren:2017:fasterrcnn}.
    }

    \label{fig:intro}
\end{figure}

The efficiency in FSOD can be measured by two metrics, \textit{i.e.}, computational complexity (per iteration FLOPs) and adaptation speed (the number of iterations required for convergence). First, computational complexity reflects per iteration computation cost of a few-shot detector during few-shot transfer (a.k.a. finetuning) or inference. The computational overheads of existing methods mainly stem from three circumstances, \textit{i.e.}, auxiliary branches, modules or inputs. Specifically, ~\cite{kang2019few,yan2019meta,fan2020few,xiao2020few} introduce an extra support \textit{branch} to extract support features as guidance for the query branch.
\cite{yang2020context} introduces a transformer \textit{module}~\cite{vaswani2017attention} to tackle the object confusion problem.
\cite{wu2020multi} proposes to add positive samples of different scales as \textit{input} to handle the scale variation problem.

Besides, adaptation speed is another significant factor that determines the efficiency of the few-shot transfer process, since the overall computational overheads are calculated as the product of per iteration FLOPs and the number of iterations required for convergence.
In fact, most existing methods generally suffer from slow adaptation issues.
Concretely, \cite{wang2020frustratingly,wu2020multi} assign randomly initialized novel class weights to the box classifier and regressor before few-shot transfer.
However, such randomly initialized weights are prone to lying far away from the optimum solution, which requires large numbers of gradient steps for convergence.
Additionally, meta-learning methods~\cite{yan2019meta,xiao2020few}, though adapting faster than existing transfer learning methods~\cite{wang2020frustratingly,wu2020multi}, still cannot achieve satisfactory adaptation speed with only a few steps, as the meta-learning principle indicates.

To address the above issues, we present an efficient pretrain-transfer framework (PTF) baseline without additional components by simply finetuning the network with our tailored adaptation strategy, which yields comparable performance with the state-of-the-art (SOTA) methods.
Upon that, we devise an initializer, namely knowledge inheritance (KI), to reliably initialize the novel class weights for the box classifier before few-shot transfer.
Our motivation is that a good initial point of the box classifier can substantially reduce few-shot optimization steps on the loss landscape and end up with fast adaptation speed.
However, the vanilla KI, which simply takes the class-wise average of feature representations followed by $\mathcal{L}$-2 normalization as the prediction for the novel class centroids (see Sec.~\ref{ki}), may undesirably introduce an inconsistency between the length of the predicted novel class weights and that of the re-used base class weights, especially when the benchmark involves large-scale vocabulary, \textit{e.g.}, LVIS of 1230 categories.
To solve this problem, we design an adaptive length re-scaling (ALR) strategy to adaptively rescale the predicted novel weights to have similar length with the base ones, which makes our approach scalable to large corpora.
Note that our proposed initializer is a free lunch that adds no overheads in both few-shot transfer and inference, since it requires no training and can be disposed after initialization.
Incorporating PTF baseline with our KI initializer not merely obtains new SOTA results across three public benchmarks, \textit{i.e.}, PASCAL VOC, COCO and LVIS,
but also shows 
$1.8\text{\textendash}100\times$ faster adaptation speed than the other methods on COCO/LVIS benchmark.

\minisection{Contributions.} We summarize our key contributions as:
\begin{itemize}
  \item A highly efficient pretrain-transfer framework (PTF) baseline without extra overheads for FSOD, which performs on par with the previous methods.
  \item Upon this baseline, we devise KI initializer to reliably initialize the novel weights for box classifier with ALR strategy to conquer the length inconsistency problem, which consistently achieves new SOTA results across three benchmarks, \textit{i.e.}, PASCAL VOC, COCO and LVIS.
  \item Our approach is highly efficient and shows
  $1.8\text{\textendash}100\times$ faster adaptation speed than the previous methods on COCO/LVIS benchmark.
  \item To our best knowledge, this is the \textit{first work} to consider the \textit{efficiency} problem in FSOD. We hope to motivate a trend toward powerful yet efficient few-shot technique development.
\end{itemize}

\section{Related Works}
\subsection{Generic Object Detection}
Deep learning based object detection can be broadly divided into two families: two-stage and one-stage detectors. Two-stage detectors~\cite{girshick2014rich,girshick2015fast,Ren:2017:fasterrcnn} first generate by the RPN module a set of region proposals, representing the regions that are shortlisted with high probability to contain objects. Then a fixed length of features will be extracted by RoI pooling for each proposal. These instance-level features will go through a transformation module, followed by a box regressor and a category classifier to obtain the final predicted bounding boxes with corresponding confidence scores. One-stage frameworks~\cite{redmon2016you,liu2016ssd} predict box coordinate and class score simultaneously based on a set of default (predefined) anchor boxes, making it amazingly fast in inference. In this paper, we choose Faster R-CNN~\cite{Ren:2017:fasterrcnn} as our basic framework for fair comparison with existing methods, though our method is model-agnostic and can be applied to any other framework.

\subsection{Few-Shot Learning}
Few-shot learning intends to empower machine vision system with the capability to quickly learn a novel concept with only limited data, and has been widely explored in the machine learning community recently, which can be mainly divided into three branches.
\romannumeral 1) Bayesian approaches. \cite{fei2006one} implements Bayesian inference to leverage prior knowledge to deal with one-shot learning problem. \cite{lake2013one} proposes a hierarchical Bayesian model based on compositionality and causality.
\romannumeral 2) Data augmentation. \cite{hariharan2017low,wang2018low} introduce image hallucination techniques to enrich the diversity of few-shot data and thereby reduce the few-shot learning difficulty.
\romannumeral 3) Meta-learning, a.k.a. ``learning to learn", explicitly tries to learn a base learner that is capable to generalize well to novel tasks by episodic training paradigm. It can be further categorized into two sub-branches as:
a) Optimization based methods MAML~\cite{finn2017model} and its variant~\cite{rajeswaran2019meta} try to explicitly train the parameters of the model such that good generalization performance could be attained with a small number of gradient steps.
b) Metric-learning based methods~\cite{snell2017prototypical,vinyals2016matching,koch2015siamese,sung2018learning} concentrate on designing good metrics to measure the similarity between support and query samples, base on which categorizing each query sample into its nearest support class in the latent space.
However, the above-mentioned methods mainly focus on few-shot classification problem and cannot be easily adapted to FSOD, since FSOD features much more complicated characteristics, such as multiple objects, complex context and occlusion.
Beyond that, few-shot learning techniques are also ubiquitously used in other fields, \textit{e.g.}, fine-grained image classification~\cite{li2020bsnet}, semantic segmentation~\cite{liu2021harmonic}, person re-identification~\cite{wu2019few}, common object localization~\cite{zhu2021few} and continual learning~\cite{chen2020did}.

\subsection{Few-Shot Object Detection}
Existing approaches in FSOD mainly follow two principles, \textit{i.e.}, meta-learning and transfer learning (a.k.a. finetuning).

\noindent\textbf{Meta-Learning.} 
\cite{fan2020few} enhances information interaction by introducing the attention-RPN and multi-relation module.
\cite{li2021transformation} introduces consistency regularization on predictions from various \textit{transformed} images to improve the generalization ability over perturbed images.
\cite{zhang2021accurate} proposes a support-query mutual guidance mechanism and a hybrid loss to enhance the discrimination ability on unseen classes.
\cite{han2021query} resorts to heterogeneous graph convolutional networks to achieve efficient message passing among all the query image regions (usually proposals) and novel class nodes.
However, distinct from our adopted \textit{generalized} FSOD~\cite{wang2020frustratingly} setting, the above methods follow the \textit{transferred} FSOD setting, where they evaluate the few-shot detector merely on novel classes.
In the generalized FSOD, \cite{kang2019few} proposes to adaptively adjust the channel-wise importance of the final \textit{global} feature maps according to different novel tasks, whereas \cite{yan2019meta} focuses on the intermediate \textit{instance-level} feature adaptation. More recently, \cite{xiao2020few} proposes a feature aggregation module to effectively combine support and query features in the late stage.
\cite{li2021beyond} proposes to optimize both feature space partition and novel class reconstruction by a class margin equilibrium approach.
\cite{hu2021dense} proposes a dense relation distillation module and an adapative context-aware feature aggregation module to exploit support information, global and local features to assist few-shot detection on novel classes.
These methods, though adopt the \textit{generalized} FSOD setting, in fact focus solely on the generalization performance to novel classes, while lacking the consideration for preventing catastrophic forgetting.

As a common problem in the meta-learning principle, all the above methods, regardless of the evaluation settings, suffer from severe efficiency issues. First, an extra support branch substantially increases the computational overheads. Even worse, the interaction module between the support and query branches further increases the overheads by $N$ times under $N$-way setting. This makes the above methods impractical when it comes to the benchmark with large numbers of classes, such as LVIS.

\noindent\textbf{Transfer learning.}
\cite{yang2020context} leverages contextual information to tackle the object confusion problem with transformer~\cite{vaswani2017attention} structure.
\cite{wu2020multi} proposes to add positive samples of different scales as input to handle the scale variation problem.
\cite{wang2020frustratingly} proposes a simple two-phase framework (TFA) without auxiliary components, which, however, takes a long journey to converge due to their inappropriate adaptation strategies.
\cite{zhu2021semantic} leverages the semantic relation between base and novel classes and introduces explicit relation reasoning into the learning of novel object detection.
\cite{zhang2021hallucination} proposes to increase the data variance for novel classes by transferring the shared within-class variation from base classes.
\cite{li2021few} introduces a few-shot classification refinement mechanism to enhance classification capability along with retreatment solutions to combat distractor samples due to incomplete annotations.
\cite{wu2021universal} enhances object features with intrinsical characteristics that are universal across different object categories to increase model generalization ability.
\cite{cao2021few} constructs a discriminative feature space for each novel class by asscociation and discrimination.
Recently, \cite{sun2021fsce} proposes a contrastive proposal encoding loss to alleviate the misclassification issues by promoting instance level intra-class compactness and interclass variance.
\cite{qiao2021defrcn} introduces gradient decoupled layer for multi-stage decoupling
and prototypical calibration block for multi-task decoupling.
Instead of introducing additional modules~\cite{yang2020context,wu2020multi,zhu2021semantic,zhang2021hallucination,li2021few,qiao2021defrcn} or loss terms~\cite{li2021few,wu2021universal,sun2021fsce,cao2021few} that increases the computational complexity on few-shot transfer or even inference as well, we propose an \textit{efficient} yet effective pretrain-transfer framework without any additional components, upon which we build the KI initializer with adaptive length re-scaling mechanism to claim the new SOTA \textit{performance} with surprisingly \textit{fast adaptation}.

\begin{figure*}[t]
	\vspace{-3mm}
	\begin{center}
		\includegraphics[width=\linewidth]{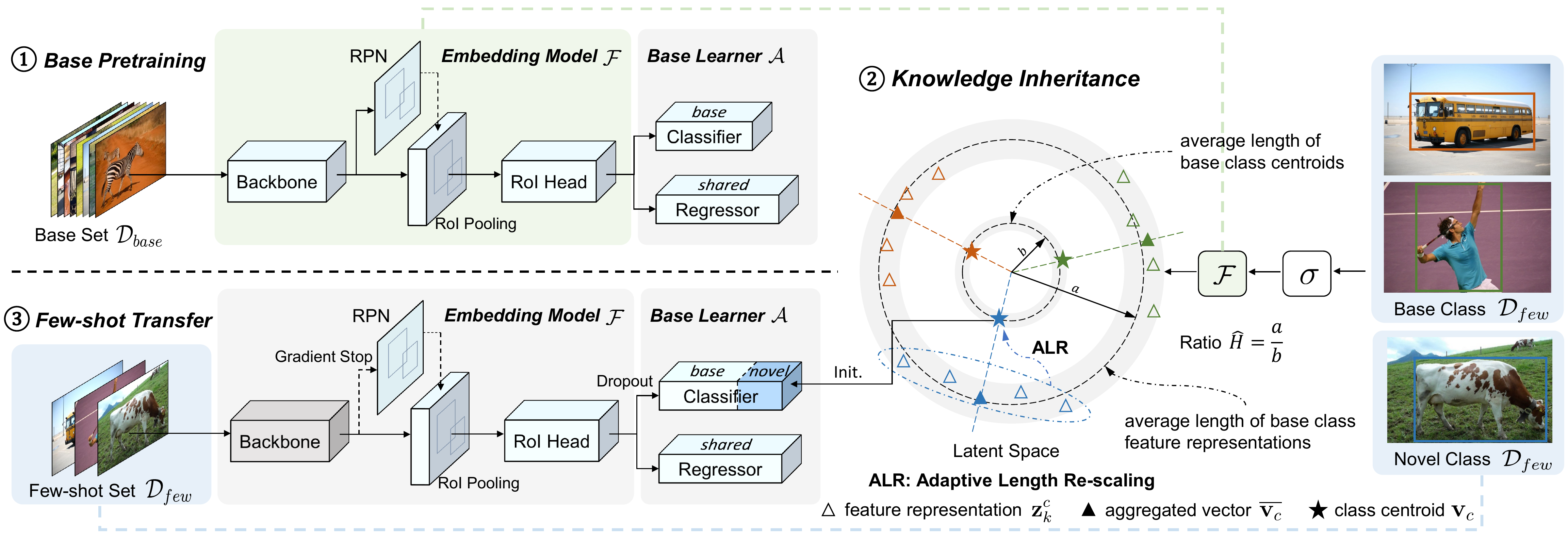}
		\caption{The overall architecture of our pretrain-transfer framework (PTF) with knowledge inheritance (KI). During base pretraining, the standard Faster R-CNN comprising of the embedding model $\mathcal{F}$ and base learner $\mathcal{A}$ is trained on $\mathcal{D}_{base}$. Then the KI initializer is utilized to initialize the novel classifier weights with the predicted novel class centroids, after which the whole detector will be finetuned on $\mathcal{D}_{few}$ to optimize both base and novel performance during few-shot transfer. In the KI initializer, the adaptive length re-scaling (ALR) strategy is adopted to alleviate the length inconsistency issue (see Sec.~\ref{ki}). Our PTF baseline excludes the KI initialization and does not update the backbone (highlighted in gray) during few-shot transfer. Details can be found in Sec.~\ref{baseline} and \ref{ki}.
		}
		\label{fig:main}
	\end{center}
\end{figure*}

\section{Method}
We first establish preliminaries and fundamental notations in Sec. \ref{formulation}. Then we present an efficient pretrain-transfer framework (PTF) baseline in Sec. \ref{baseline}. Upon this baseline, we propose an initializer namely knowledge inheritance (KI) to initialize the novel weights for the box classifier in Sec. \ref{ki}, which facilitates the knowledge transfer process and boosts the adaptation speed. 

\subsection{Preliminaries}\label{formulation}
FSOD aims at effectively generalizing a detector pretrained on large-scale dataset to address unseen novel tasks with only limited training samples. Concretely, there are two disjoint datasets 
$\mathcal{D}_{base}=\left\{\left(\mathbf{x}_{i}, y_{i}\right)\right\}_{i=1}^{I}$ and 
$\mathcal{D}_{novel}=\left\{\left(\mathbf{x}_{j}, y_{j}\right)\right\}_{j=1}^{J}$ ($J\ll I$),
where $\mathbf{x}_i / \mathbf{x}_j$ is the input image and $y_i\in C_{base} / y_j\in C_{novel}$ the corresponding class label and box coordinates. Note that $C_{base}$ and $C_{novel}$ are defined to be non-overlapped ($C_{base}\cap C_{novel}=\emptyset$) in order to effectively evaluate the generalization capacity of few-shot detectors. $\mathcal{D}_{base}$ generally contains large numbers of annotated images while $\mathcal{D}_{novel}$ has only $K$-shot (typically $K\leq 10$) instances for each category.
Additionally, few-shot object detectors are evaluated on a test set $\mathcal{D}^{test}\sim C_{base}\cup C_{novel}$ encompassing both base and novel categories. 
Therefore, a class-balanced dataset $D_{few}$ for few-shot transfer is constructed by combining $D_{novel}$ with the sub-sampled base dataset $\hat{D}_{base}$ that contains $K$-shot instances per base category from $D_{base}$. The objective is to optimize the detection performance measured by average precision (AP) of the novel classes as well as the base classes, which we refer to as generalized FSOD.
Finally, unlike the standard $N$-way $K$-shot (typically $N=1,5$) scheme in few-shot classification, the evaluation protocol~\cite{kang2019few} in FSOD follows a more challenging setting, \textit{i.e.}, full-way ($N=\left \| C_{base}\cup C_{novel} \right \|$) $K$-shot scheme.

\subsection{Pretrain-Transfer Framework}\label{baseline}

Our pretrain-transfer framework (PTF) consists of two phases, \textit{i.e.}, the \textit{base pretraining} phase to establish prior knowledge from large-scale dataset $D_{base}$ and the \textit{few-shot transfer} phase to effectively transfer the learned prior knowledge to facilitate unseen few-shot detection tasks.
We adopt Faster R-CNN~\cite{Ren:2017:fasterrcnn}, the same framework with previous SOTA methods~\cite{yan2019meta,wang2020frustratingly,wu2020multi,xiao2020few,sun2021fsce,qiao2021defrcn,wu2021universal,zhu2021semantic}, as our base detector for fair comparison, but do note that our method is model-agnostic. For clarity, we divide the overall architecture into two parts, \textit{i.e.}, the embedding model $\mathcal{F}$ for feature extraction and the base learner $\mathcal{A}$ for final prediction. As shown in Fig.~\ref{fig:main}, the embedding model comprises a backbone $\phi$, a region proposal network (RPN) $\gamma$  and 
a proposal-level feature extractor namely RoI head $\psi$. The base learner includes a classifier $\omega^{cls}$ to classify object categories and a box regressor $\omega^{loc}$ to locate the coordinates.

\minisection{Base pretraining.} For the sake of efficiency and conciseness, we establish prior knowledge simply by pretraining a standard Faster R-CNN model on $D_{base}$ without bells and whistles, which can be given as:
\begin{equation}
\theta,\omega=\underset{\theta,\omega}{\arg \min } 
\mathcal{L}^{S} \left( \mathcal{A} \left( \mathcal{F} \left(\mathcal{D}_{base} ; \theta \right); \omega \right) \right)+ \mathcal{R}(\theta,\omega),
\label{eq:std-train}
\end{equation}
where $\theta=\{\phi,\gamma,\psi\}$, $\omega=\{\omega^{cls},\omega^{loc}\}$, $\mathcal{L}^S=\mathcal{L}_{\text{rpn}} + \mathcal{L}_{\text{cls}} + \mathcal{L}_{\text{loc}}$ is the standard loss function adopted in \cite{Ren:2017:fasterrcnn} and $\mathcal{R}$ the regularization term.
Note that distinct from \cite{wang2020frustratingly}, we set the box regressor $\omega^{cls}$ as class-agnostic so that it can be directly re-used as a good initialization for few-shot transfer.

\minisection{Few-shot transfer.} In this phase, the objective is to sufficiently leverage those helpful prior knowledge contained in the pretrained base model to assist with novel tasks while maintaining \textit{high efficiency}.
Following \cite{sun2021fsce}, apart from the base learner, we also update the RPN \hl{(maximum number of proposals kept after NMS doubled)} and RoI head during few-shot transfer.
\hl{We notice that finetuning the RPN can provide more RoI proposals for unseen novel objects and  improve the average recall (AR) for novel classes.
Besides, updating the RoI head not only provides better feature embedding for novel classes but, more importantly, substantially boosts the adaptation speed, which perfectly coincides with our objective --- \textit{efficiency}.}
Distinct from \cite{sun2021fsce}, we do not update the backbone for the PTF baseline as it increases the computational overhead while also harming the base performance. Quantitative analysis can be found in Sec.~\ref{exp:ablation_vis}.
In brief, we perform the few-shot transfer as:
\begin{equation}
\resizebox{0.9\linewidth}{!}{$
\theta',\omega=\underset{\theta',\omega}{\arg \min } 
\mathcal{L}^{S} \left( \mathcal{A} \left( \mathcal{F} \left(\mathcal{D}_{base} ; \theta' \right); \omega \right) \right)+ \mathcal{R}(\theta',\omega),
$}
\label{eq:our-ft}
\end{equation}
where $\theta'=\{\gamma,\psi\}$, $\omega=\{\omega^{cls},\omega^{loc}\}$, $\mathcal{L}^S$ and $\mathcal{R}$ are the same as Eq.~\ref{eq:std-train}.

Note that the box regressor is directly initialized with the pretrained weights and the classifier is extended with randomly initialized weights for the novel classes.
We also investigate two instantiations of the classifier, \textit{i.e.}, the linear classifier and cosine similarity based classifier. We keep the linear classifier as it yields better results.

\begin{figure}[t]
	\vspace{-2mm}
    \centering
	\includegraphics[width=0.9\linewidth]{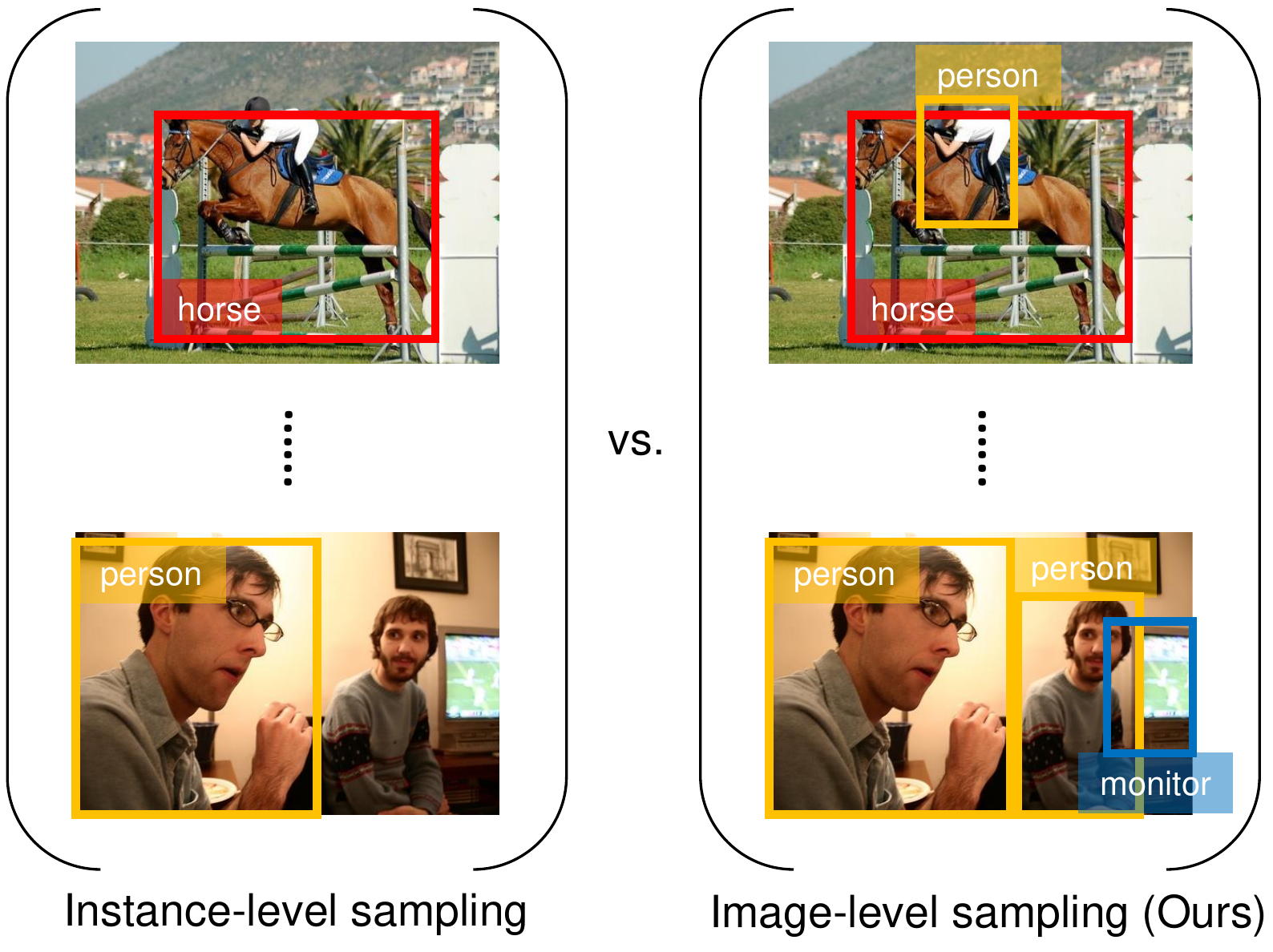}
	\caption{\hl{Illustration of different batch sampling strategies.}}
	\label{fig:batch_sampling}
\end{figure}

\minisection{Batch sampling.} Given few-shot dataset $D_{few}$, \hl{existing methods~\cite{wang2020frustratingly,sun2021fsce}} form a batch by instance-level sampling with each image associated with only one selected instance (ignoring the rest instances). In this way, \hl{as illustrated in Fig.~\ref{fig:batch_sampling}}, the ignored objects \hl{(\textit{e.g.}, person, monitor)} with corresponding annotations removed will be treated as background during loss computation and therefore harm the performance.
We address this issue by image-level sampling with all available instance annotations within the selected images retained to form a batch. Note that the same $D_{few}$ is used except for the different batching schemes.

\minisection{Scaled learning rate.} With the aim of fast adaptation, we use 10 times larger global learning rate than previous methods~\cite{wang2020frustratingly,sun2021fsce}. However, we find that using a large global learning rate for RoI head will undesirably undermine the base performance due to overfitting given the scarce few-shot data $D_{few}$. This is because RoI head consists of larger numbers of weights compared with the base learner and RPN. As a result, we apply the scaled learning rate strategy with each component assigned with a specific scaling factor $\alpha(\cdot)$, where $(\cdot)\in \{\phi,\gamma,\psi,\omega\}$. We set the scaling factor as 1x for base learner and RPN while 0.5x for RoI head. The actual learning rate $lr(\cdot)$ for each component will be the global learning rate $lr$ multiplied by the corresponding scaling factor $\alpha(\cdot)$, \textit{e.g.}, $lr(\psi)=lr\times \alpha(\psi)$ for the RoI head.

\minisection{Dropout strategy.} During few-shot transfer, it is common that a model learns too ``aggressively'' on the few-shot dataset $D_{few}$ such that it drastically forgets the base knowledge while not performing preferably on the novel tasks as well due to severe overfitting. To alleviate this issue, we propose to adopt the dropout strategy as an additional regularization during few-shot adaptation.
Specifically, we add a dropout layer before the final classifier with a dropout rate of 0.8, as shown in Fig.~\ref{fig:main}. 
We empirically find that this simple strategy effectively benefits both base and novel performance.

\subsection{Knowledge Inheritance}\label{ki}
In our PTF baseline, the classifier $\omega^{cls}$ needs to be randomly initialized for the novel classes while can be reliably initialized with the pretrained weights for the base classes. This inconsistent initial point tends to fluctuate the optimization steps and slow down the adaptation speed.
To alleviate this issue, we devise an initializer, namely knowledge inheritance (KI), to reliably initialize the novel weights for the classifier \textit{without network training}, which facilitates the knowledge transfer process and boosts the adaptation speed. Note that we unfreeze the backbone with a learning rate scaling factor 0.01x and stop the gradient from the RPN module when equipped with KI.

Our key insight is that the feature representations of the novel objects, encoded by the base embedding model $\mathcal{F}$, contain abundant prior knowledge and can be leveraged to predict the \hl{novel class centroids $\{\mathbf{v}_c|c\in C_{novel}\}$}, where $\mathbf{v}_c \in \mathbb{R}^d$ is the per-class weight vector.
Concretely, given the pretrained base detector, its embedding model $\mathcal{F}$ is utilized as a feature extractor to map each few-shot image-label pair $(\mathbf{x}_k^c, y_k^c) \in D_{few}$ containing the $k$-th shot instance of category $c$ into the latent space and extract the corresponding instance-level feature representation $\mathbf{z}_k^c=\mathcal{F}\left(\sigma \left(\mathbf{x}_k^c\right)\right)\in \mathbb{R}^d$ (depicted with hollow triangle in Fig.~\ref{fig:main}), 
where $\sigma(\cdot)$ denotes the same data augmentation strategies (\textit{i.e.}, random scaling and horizontal flip) as base pretraining.
Here the RoI pooling in $\mathcal{F}$ takes as input the ground truth boxes rather than the predicted RoI proposals. Then the remaining problem is how to effectively aggregate the feature representations $\mathbf{z}_k^c$ to predict each \hl{novel class centroid $\mathbf{v}_c$}.

Firstly, we take the class-wise average over multiple shots each with $R$ randomly augmented views to represent the direction of each \hl{novel class centroid} with an aggregated vector
\hl{$\overline{\mathbf{v}_c}=\frac{1}{K\times R}\sum_{k=1}^{K}\sum_{r=1}^{R} (\mathbf{z}_k^c)_r$},
where $(\cdot)_r$ denotes the $r$-th augmented view and $\overline{\mathbf{v}_c} \in \mathbb{R}^d$.
This aggregation strategy is motivated by our \textit{first observation} that object feature representations (hollow triangles) from the same category tend to form a cluster in the same direction with their corresponding class centroid (solid 5-point star) as shown in Fig.~\ref{fig:main}. Then we need to rescale the aggregated vector $\overline{\mathbf{v}_c}$ to appropriate length (euclidean norm) so as to avoid exploding gradients.
Inspired by \cite{qi2018low}, we apply the $\mathcal{L}$-2 normalization to each aggregated vector \hl{$\overline{\mathbf{v}_c}$} and obtain the corresponding \hl{novel class centroid $\mathbf{v}_c=\mathcal{N}\left(\overline{\mathbf{v}_c}\right) \in \mathbb{R}^d$}.
However, this re-scaling approach assumes that all novel class centroids should be in unit length, which may cause an inconsistency between the length of the re-used base class centroids and the predicted novel class centroids. It will drastically degrade the base performance as discussed in Sec. \ref{exp:ablation_vis}.

\begin{figure}[t]
	\vspace{-2mm}
    \centering
	\includegraphics[width=0.9\linewidth]{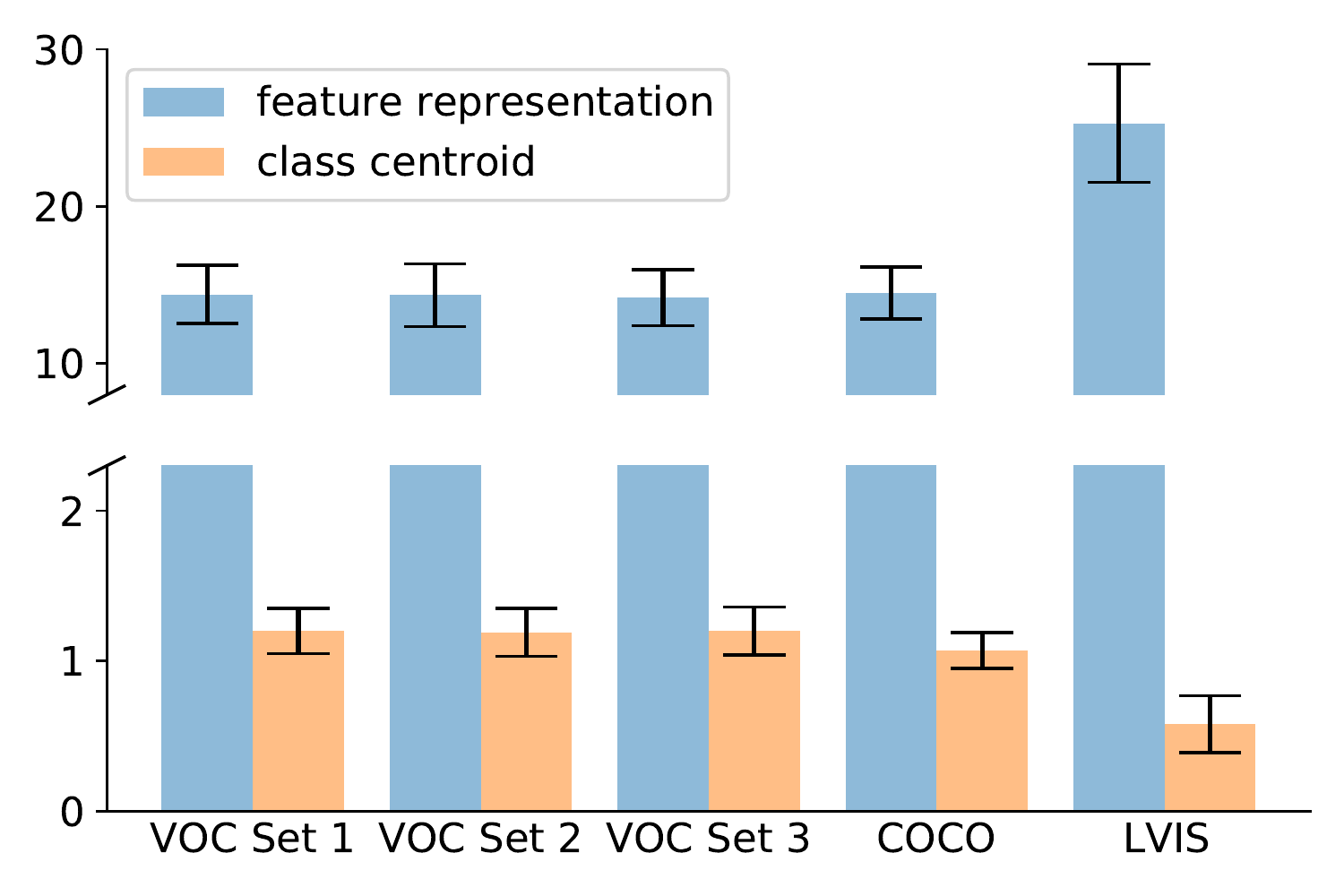}
	\caption{\hl{Supporting evidence for the second observation --- hyper-sphere. The length of either base class feature representations or class centroids consistently has \textit{low} standard deviation across different benchmarks.}}
	\label{fig:observation2}
\end{figure}

\begin{table*}[ht]
  \centering
  \footnotesize
  \setlength{\tabcolsep}{0.4em}
  \adjustbox{width=\linewidth}{
    \begin{tabular}{l|c|ccccc|ccccc|ccccc}
    \toprule
    \multirow{2}{*}{Method / Shot} & \multirow{2}{*}{Backbone} & \multicolumn{5}{c|}{Novel Set 1} & \multicolumn{5}{c|}{Novel Set 2} & \multicolumn{5}{c}{Novel Set 3} \\ 
          &       & 1     & 2     & 3     & 5     & 10    & 1     & 2     & 3     & 5     & 10    & 1     & 2     & 3     & 5     & 10 \\ \midrule
    YOLO-ft-full~\cite{kang2019few}  & \multirow{3}{*}{YOLOv2} & 6.6   & 10.7  & 12.5 & 24.8 & 38.6 & 12.5  & 4.2   & 11.6 & 16.1 & 33.9 & 13.0  & 15.9  & 15.0 & 32.2 & 38.4 \\ 
    FSRW~\cite{kang2019few}  &  & 14.8  & 15.5  & 26.7  & 33.9  & 47.2  & 15.7  & 15.3  & 22.7  & 30.1  & 40.5  & 21.3  & 25.6  & 28.4  & 42.8  & 45.9 \\
    MetaDet~\cite{wang2019meta} &       & 17.1  & 19.1  & 28.9  & 35.0  & 48.8  & 18.2  & 20.6  & 25.9  & 30.6  & 41.5  & 20.1  & 22.3  & 27.9  & 41.9  & 42.9 \\ \midrule
    FRCN+ft-full~\cite{wang2020frustratingly} &  & 15.2 & 20.3 & 29.0 & 40.1 & 45.5 & 13.4 & 20.6 & 28.6 & 32.4 & 38.8 & 19.6 & 20.8 & 28.7 & 42.2 & 42.1 \\
    Meta R-CNN~\cite{yan2019meta} &  & 19.9  & 25.5  & 35.0  & 45.7  & 51.5  & 10.4  & 19.4  & 29.6  & 34.8  & 45.4  & 14.3  & 18.2  & 27.5  & 41.2  & 48.1 \\
    FSDetView~\cite{xiao2020few} &       & 24.2  & 35.3  & 42.2  & 49.1  & 57.4  & 21.6  & 24.6  & 31.9  & 37.0  & 45.7  & 21.2  & 30.0  & 37.2  & 43.8  & 49.6 \\
    MPSR$\dagger$~\cite{wu2020multi}  &       & 36.5  & 41.7  & 48.9  & 54.5  & 60.7  & 25.4 & 27.2  & 39.5  & 41.1 & 46.2 & 35.3 & 41.2  & 44.5 & 49.6  & 50.4 \\
    TFA w/ fc~\cite{wang2020frustratingly} &  & 36.8  & 29.1  & 43.6  & 55.7  & 57.0  & 18.2  & 29.0  & 33.4  & 35.5  & 39.0  & 27.7  & 33.6  & 42.5  & 48.7  & 50.2 \\
    TFA w/ cos~\cite{wang2020frustratingly} &   & 39.8  & 36.1  & 44.7  & 55.7  & 56.0  & 23.5  & 26.9  & 34.1  & 35.1  & 39.1  & 30.8  & 34.8  & 42.8  & 49.5  & 49.8 \\
    FSCE$\dagger$~\cite{sun2021fsce} &  & 41.1 & 41.9 & 49.6 & 57.7 & 61.5 & 21.7 & 28.8 & \underline{42.9} & \underline{44.2} & \underline{50.6} & 37.3 & 45.7 & 46.2 & \underline{55.3} & \underline{58.4} \\
    SRR-FSD~\cite{zhu2021semantic} &  & \underline{47.8} & 50.5 & 51.3 & 55.2 & 56.8 & \textbf{32.5} & \textbf{35.3} & 39.1 & 40.8 & 43.8 & \underline{40.1} & 41.5 & 44.3 & 46.9 & 46.4 \\
    $\text{FSOD}^{up}$~\cite{wu2021universal} &  & 43.8 & 47.8 & 50.3 & 55.4 & 61.7 & 31.2 & 30.5 & 41.2 & 42.2 & 48.3 & 35.5 & 39.7 & 43.9 & 50.6 & 53.5\\
    \rowcolor{Gray} PTF (Ours)   &  & 47.3 & \underline{52.8} & \underline{52.2} & \underline{59.0} & \underline{61.9} & 24.0 & 31.7 & 41.0 & 41.0 & 41.4 & 38.5 & \underline{46.0} & \underline{46.9} & 52.1 & 53.7 \\ 
    \rowcolor{Gray} PTF+KI (Ours) &  \multirow{-11}{*}{\shortstack{FRCN w/\\ R-101 FPN}} & \textbf{48.1} & \textbf{53.0}  & \textbf{54.1} & \textbf{61.2}  & \textbf{64.1} & \underline{28.5}  & \underline{32.7} & \textbf{44.2} & \textbf{44.6} & \textbf{51.6}  & \textbf{41.1}  & \textbf{50.1} & \textbf{50.3} & \textbf{57.5} & \textbf{59.1} \\ \midrule
    DeFRCN$\dagger$~\cite{qiao2021defrcn} &  & 52.3 & 57.9 & 61.9 & 65.9 & 66.8 & 34.8 & 33.6 & 49.5 & 52.2 & 51.9 & 48.4 & 52.9 & 54.1 & 60.3 & 61.6 \\
    \rowcolor{Gray} PTF+KI (Ours) &  \multirow{-2}{*}{\shortstack{FRCN w/\\ R-101 C4}} & \textbf{57.0} & \textbf{62.3} & \textbf{63.3} & \textbf{66.2} & \textbf{67.6} & \textbf{42.8} & \textbf{44.9} & \textbf{50.5} & \textbf{52.3} & \textbf{52.2} & \textbf{50.8} & \textbf{56.9} & \textbf{58.5} & \textbf{62.1} & \textbf{63.1} \\
    
    \bottomrule
    \end{tabular}}
    \caption{Few-shot detection performance on PASCAL VOC. We report the novel AP50 of three different base/novel splits, following the same setting with \cite{kang2019few}.
    $\dagger$ denotes that the results are reproduced with the official codes. The best and second best numbers are highlighted in bold and underline respectively.
    }
  \label{tab:main_voc}
\end{table*}

\minisection{Adaptive length re-scaling.}
To alleviate the inconsistency problem, we propose an adaptive length re-scaling (ALR) strategy to adaptively rescale the aggregated vector \hl{$\overline{\mathbf{v}_c}\ (c\in C_{novel})$} based on the length of the base class centroids so that they can be compatible with each other.
This design is inspired by our \textit{second observation} that feature representations and class centroids tend to approximately lie on two hyper-spheres with different radii respectively (dashed circles in Fig.~\ref{fig:main}). It suggests that there exists a ratio $H$ between the averaged length of the feature representations and that of the class centroids, which can be utilized to rescale the aggregated vectors \hl{$\overline{\mathbf{v}_c}\ (c\in C_{novel})$} to be on the same hyper-sphere with the base class centroids. Specifically, we first estimate the ratio $H$ from the observation of the base classes, which can be given as:
\begin{equation}
\hat{H} = \frac{\mathbb{E}_{\hl{c\in C_{base}}}\left[\left \| \frac{1}{K\times R}\sum_{k=1}^{K}\sum_{r=1}^{R} (\mathbf{z}_k^c)_r \right \|_2\right]}{\mathbb{E}_{\hl{c\in C_{base}}}\left[\left \| \mathbf{v}_c \right \|_2\right]},
\end{equation}
where $(\cdot)_r$ denotes the $r$-th augmented view. We empirically find that multiple augmented views can enhance the robustness of the initialization quality and hyperparameter $R$ is set to 10 in our experiments.
Then we apply the estimated ratio $\hat{H}$ to re-scaling the aggregated vectors to obtain the final \hl{novel class centroids as $\mathbf{v}_c=\overline{\mathbf{v}_c}\,/\,\hat{H}$}. Finally, we initialize the classifier weights $\omega^{cls}_c\in \mathbb{R}^d$ for each \hl{novel class $c$} simply with the obtained corresponding class centroid $\mathbf{v}_c\in \mathbb{R}^d$, while the bias term is initialized with zero by default.

\minisection{Supporting evidence for two observations.}
To be convincing, we provide supporting evidence for our two observations that incubate the design of our KI initializer. To verify the first observation, we normalize the base class (VOC split 1) feature representations and class centroids to unit length and then visualize them with t-SNE~\cite{van2008visualizing} \hl{(refer to the appendix)}.
For the second observation, we show the average length of base class feature representations and class centroids respectively across different benchmarks in Fig.~\ref{fig:observation2}. The consistently low standard deviation of the length of either feature representations or class centroids across different benchmarks proves our second observation that either feature representations or class centroids tend to lie on certain hyper-sphere regardless of their categories.

\section{Experiments}
In this section, we first describe the benchmarks and setups in Sec. \ref{exp:benchmark}. Then we comprehensively verify the superiority of our proposed method in terms of \textit{performance} (Sec. \ref{exp:performance}) and \textit{efficiency} (Sec. \ref{exp:efficiency}). Note that efficiency comprises computational complexity and adaptation speed, and has barely been discussed in previous works. Finally, we present thorough ablation studies and visualizations in Sec. \ref{exp:ablation_vis}.

\subsection{Benchmarks and Setups}\label{exp:benchmark}

We conduct extensive comparisons with existing methods
across three different benchmarks with hierarchical difficulty, \textit{i.e.} PASCAL VOC (preliminary), COCO (moderate) and LVIS (difficult).
Concretely, following \cite{kang2019few}, three different base/novel class splits are considered for PASCAL VOC, where there are 15 base classes and 5 novel classes each with $K=1,2,3,5,10$ instance(s) for few-shot transfer. For consistency, we adopt the same few-shot dataset $D_{few}$ with \cite{kang2019few} and use VOC 2007 test set for evaluation.
For COCO, we treat the 20 classes that are overlapped with PASCAL VOC as novel classes while the rest as base classes.
We follow recent works \cite{wang2020frustratingly,xiao2020few} to report the results averaged over 10 random runs on COCO val2017 set with $K=10,30$.
For LVIS, we group the frequent and common classes as base split, and the rare classes as novel split. Following \cite{wang2020frustratingly}, we construct a balanced few-shot dataset by sampling 10 instances for each base class while including all for novel classes. All results are reported with the standard evaluation protocol in each benchmark.
Finally, we notice that \texttt{DeFRCN}~\cite{qiao2021defrcn} uses a different base detector architecture Faster R-CNN with ResNet101-C4 from the prevailing one, \textit{i.e.}, Faster R-CNN with ResNet101-FPN. For fairness, we build our \texttt{PTF+KI} upon \texttt{DeFRCN} when compared with it. Note that the post-processing PCB module proposed by \texttt{DeFRCN} is not applied in all experiments for the sake of efficiency.

\minisection{Implementation details.} We adopt Faster R-CNN~\cite{Ren:2017:fasterrcnn} as our base detector with ResNet101-FPN~\cite{he2016deep,lin2017feature} and ResNet101-C4 as backbone. For training details, we follow the default hyperparameters in detectron2~\cite{wu2019detectron2}, where models are trained with momentum (0.9) SGD optimizer, batch size of 16, and weight decay of 1e-4. An initial learning rate 0.02 is used during base pretraining and 0.02 for PASCAL VOC or 0.01 for COCO and LVIS during few-shot transfer. For 10-shot setting, 500 / 1000 / 3000 iterations are taken during few-shot transfer for PASCAL VOC / COCO / LVIS respectively.
All experiments are run on four RTX 2080 Ti GPUs.

\begin{table}[t]
\centering
\footnotesize
\setlength{\tabcolsep}{0.5em}
\adjustbox{width=\linewidth}{
\begin{tabular}{l|c|cc|cc}
\toprule
\multirow{2}{*}{Method/Shot} & \multirow{2}{*}{Backbone} & \multicolumn{2}{c|}{3} & \multicolumn{2}{c}{10}  \\
& & bAP & nAP & bAP & nAP \\ \midrule
FRCN+ft-full~\cite{wang2020frustratingly} & & 66.1 & 29.0 & 66.0 & 45.5\\
Meta R-CNN~\cite{yan2019meta} & & 64.8 & 35.0 & 67.9 & 51.5 \\
FSDetView~\cite{xiao2020few} & & 65.9 & 42.2 & 69.1 & 57.4\\
MPSR$\dagger$~\cite{wu2020multi} & & 67.6 & 48.9 & 71.3 & 60.7\\
TFA w/ cos~\cite{wang2020frustratingly} & & \textbf{79.1} & 44.7  & 78.4 & 56.0 \\
FSCE$\dagger$~\cite{sun2021fsce} &  & 74.3 & 49.6 & 76.7 & 61.5 \\
SRR-FSD~\cite{zhu2021semantic} &  & 78.2 & 51.3 & 78.2 & 56.8 \\
$\text{FSOD}^{up}$~\cite{wu2021universal} &  & 66.3 & 50.3 & 69.7 & 61.7 \\
Train base only & & 81.0 & - & 81.0 & - \\
\rowcolor{Gray} PTF (Ours) & & \underline{78.9} & \underline{52.2} & \textbf{79.3} & \underline{61.9}\\
\rowcolor{Gray} PTF+KI (Ours) & \multirow{-11}{*}{\shortstack{FRCN w/\\ R-101 FPN}} & 78.3 & \textbf{54.1} & \underline{78.7} & \textbf{64.1}\\ \midrule
DeFRCN$\dagger$~\cite{qiao2021defrcn} & & 76.1 & 61.9 & 77.0 & 66.8 \\
Train base only &  & 80.3 & - & 80.3 & - \\
\rowcolor{Gray} PTF+KI (Ours) & \multirow{-3}{*}{\shortstack{FRCN w/\\ R-101 C4}} & \textbf{78.0} & \textbf{63.3} & \textbf{78.5} & \textbf{67.6}\\
\bottomrule
\end{tabular}}
\caption{Few-shot detection performance (AP50) for the base (bAP) and novel (nAP) classes on the first novel set of PASCAL VOC. Our approach outperforms existing methods on both base and novel AP and does not suffer from catastrophic forgetting like \cite{yan2019meta,wu2020multi,xiao2020few,wu2021universal}. $\dagger$ denotes that the results are reproduced with the official codes.}
\label{tab:voc_base}
\end{table}

\subsection{Performance}\label{exp:performance}

\begin{table*}[ht]
  \centering
  \footnotesize
  \setlength{\tabcolsep}{0.7em}
    \renewcommand{\arraystretch}{0.8}
    \adjustbox{width=\linewidth}{
        \begin{tabular}{cl|cccccc|cccccc}
        \toprule
        \multirow{2}{*}{Shot} & \multirow{2}{*}{Method} & \multicolumn{6}{c|}{Average Precision} & \multicolumn{6}{c}{Average Recall} \\
        & & 0.5:0.95  & 0.5  & 0.75 & S & M & L & 1  & 10 & 100 & S & M & L \\ \midrule
        \multirow{12}[0]{*}{10} & FSRW~\cite{kang2019few}  & 5.6   & 12.3  & 4.6   & 0.9   & 3.5   & 10.5  & 10.1  & 14.3  & 14.4  & 1.5   & 8.4   & 28.2 \\
              & MetaDet~\cite{wang2019meta} & 7.1   & 14.6  & 6.1   & 1.0   & 4.1   & 12.2  & 11.9  & 15.1  & 15.5  & 1.7   & 9.7   & 30.1 \\
              & Meta R-CNN~\cite{yan2019meta} & 8.7   & 19.1  & 6.6   & 2.3   & 7.7   & 14.0  & 12.6  & 17.8  & 17.9  & 7.8   & 15.6  & 27.2 \\
              & FSDetView$\dagger$~\cite{xiao2020few} & 10.3  & \textbf{25.1} & 6.1   & 3.5   & 11.3  & 14.6  & \underline{18.1}  & 23.8  & 23.8  & 9.3   & 24.9  & 31.1 \\
              & MPSR~\cite{wu2020multi}  & 9.8   & 17.9  & 9.7   & 3.3   & 9.2   & 16.1  & 15.7  & 21.2  & 21.2  & 4.6   & 19.6  & 34.3 \\
              & TFA* w/ fc~\cite{wang2020frustratingly} & 9.1   & 17.3  & 8.5   & 3.6     & 8.1     & 14.3     & 14.0     & 20.0     & 20.1     & 6.2     & 16.3     & 31.4 \\
              & TFA* w/ cos~\cite{wang2020frustratingly} & 9.1   & 17.1  & 8.8   & 3.7     & 8.0     & 14.3     & 13.9     & 19.9     & 20.1     & 7.3     & 16.8     & 29.6 \\
              & FSCE*$\dagger$~\cite{sun2021fsce} & 11.4 & 23.3 & 10.1 & 4.5 & 10.8 & \textbf{18.7} & - & - & - & - & - & - \\
              & SRR-FSD~\cite{zhu2021semantic} & 11.3 & 23.0 & 9.8 & - & - & - & - & - & - & - & - & - \\
              & $\text{FSOD}^{up}$~\cite{wu2021universal} & 11.0 & - & 10.7 & 4.5 & 11.2 & 17.3 & - & - & - & - & - & - \\
              \rowcolor{Gray} & PTF* (Ours) & \underline{11.7} & 22.6 & \underline{10.9} & \underline{5.1} & \underline{12.2} & 16.5 & \underline{18.1} & \underline{32.2} & \underline{33.5} & \underline{16.1} & \underline{33.8} & \underline{46.5} \\
              \rowcolor{Gray} & PTF+KI* (Ours) & \textbf{13.0} & \underline{24.0} & \textbf{12.6} & \textbf{6.0} & \textbf{13.1} & \underline{18.4} & \textbf{19.7} & \textbf{36.0} & \textbf{37.9} & \textbf{20.4} & \textbf{38.2} & \textbf{51.3} \\\midrule
        \multirow{12}[0]{*}{30} & FSRW~\cite{kang2019few}  & 9.1   & 19.0  & 7.6   & 0.8   & 4.9   & 16.8  & 13.2  & 17.7  & 17.8  & 1.5   & 10.4  & 33.5 \\
              & MetaDet~\cite{wang2019meta} & 11.3  & 21.7  & 8.1   & 1.1   & 6.2   & 17.3  & 14.5  & 18.9  & 19.2  & 1.8   & 11.1  & 34.4 \\
              & Meta R-CNN~\cite{yan2019meta} & 12.4  & 25.3  & 10.8  & 2.8   & 11.6  & 19.0  & 15.0  & 21.4  & 21.7  & 8.6   & 20.0  & 32.1 \\
              & FSDetView$\dagger$~\cite{xiao2020few} & 14.2  & \textbf{31.4} & 10.3  & 4.7   & 15.0  & 21.5 & \textbf{22.1}  & 28.8  & 28.8  & 12.1 & 29.9 & 39.4 \\
              & MPSR~\cite{wu2020multi}  & 14.1  & 25.4  & 14.2  & 4.0   & 12.9  & 23.0  & 17.7  & 24.2  & 24.3  & 5.5   & 21.0  & 39.3 \\
              & TFA* w/ fc~\cite{wang2020frustratingly} & 12.0  & 22.2  & 11.8  & 4.4 & 11.0 & 18.7 & 16.6     & 23.8     & 23.9     & 7.1     & 20.0     & 36.5 \\
              & TFA* w/ cos~\cite{wang2020frustratingly} & 12.1  & 22.0  & 12.0  & 4.7     & 10.8     & 18.6     & 16.3     & 23.2     & 23.3     & 7.4     & 19.5     & 34.6 \\
              & FSCE*$\dagger$~\cite{sun2021fsce} & 15.8 & 29.9 & 14.7 & 6.1 & \underline{16.5} & \underline{24.4} & - & - & - & - & - & - \\
              & SRR-FSD~\cite{zhu2021semantic} & 14.7 & 29.2 & 13.5 & - & - & - & - & - & - & - & - & - \\
              & $\text{FSOD}^{up}$~\cite{wu2021universal} & 15.6 & - & \underline{15.7} & 4.7 & 15.1 & \textbf{25.1} & - & - & - & - & - & - \\
              \rowcolor{Gray} & PTF* (Ours)  & \underline{15.7} & 30.4 & 14.4 & \underline{7.3} & 16.3 & 21.4 & 20.2 & \underline{36.6} & \underline{38.8} & \underline{20.9} & \underline{39.5} & \underline{51.7} \\
              \rowcolor{Gray} & PTF+KI* (Ours) & \textbf{16.8} & \underline{30.9} & \textbf{16.2} & \textbf{8.0} & \textbf{17.1} & 23.0 & \underline{21.6} & \textbf{39.1} & \textbf{41.5} & \textbf{23.6} & \textbf{42.0} & \textbf{55.2} \\
        \bottomrule
        \end{tabular}
    }
  \caption{Generalized few-shot detection performance for the novel classes on COCO (\textbf{FRCN w/ R-101 FPN}). Our PTF+KI outperforms the other methods across different shots by a considerable margin, especially in terms of average recall. * indicates that the results are averaged over 10 random seeds. 
  $\dagger$ denotes that the results are reproduced with the official codes. The best and second best numbers are highlighted in bold and underline respectively.}
  \label{tab:coco_main_fpn}
\end{table*}

\begin{table*}[ht]
  \centering
  \footnotesize
  \setlength{\tabcolsep}{0.8em}
    \renewcommand{\arraystretch}{0.8}
    \adjustbox{width=\linewidth}{
        \begin{tabular}{cl|cccccc|cccccc}
        \toprule
        \multirow{2}{*}{Shot} & \multirow{2}{*}{Method} & \multicolumn{6}{c|}{Average Precision} & \multicolumn{6}{c}{Average Recall} \\
        & & 0.5:0.95  & 0.5  & 0.75 & S & M & L & 1  & 10 & 100 & S & M & L \\
        \midrule
        & DeFRCN*$\dagger$~\cite{qiao2021defrcn}  & 16.0 & 29.0 & 15.9 & 6.1 & 16.1 & 22.7 & 19.9 & 30.6 & 31.1 & 12.9 & 30.0 & 43.7 \\
        \rowcolor{Gray} \multirow{-2}[0]{*}{10} & PTF+KI* (Ours) & \textbf{16.9} & \textbf{30.1} & \textbf{16.7} & \textbf{7.2} & \textbf{17.7} & \textbf{24.2} & \textbf{21.6} & \textbf{36.7} & \textbf{38.1} & \textbf{20.5} & \textbf{38.4} & \textbf{52.4} \\\midrule
        
        & DeFRCN*$\dagger$~\cite{qiao2021defrcn}  & 20.0 & 35.9 & 19.7 & 8.3 & 20.5 & 28.1 & 22.4 & 35.2 & 36.1 & 16.9 & 35.4 & 49.8 \\
        \rowcolor{Gray} \multirow{-2}[0]{*}{30} & PTF+KI* (Ours) & \textbf{20.7} & \textbf{37.2} & \textbf{20.4} & \textbf{9.0} & \textbf{21.5} & \textbf{29.4} & \textbf{23.5} & \textbf{40.1} & \textbf{42.3} & \textbf{24.1} & \textbf{43.0} & \textbf{57.4} \\
    \bottomrule
    \end{tabular}
    }
  \caption{Generalized few-shot detection performance for the novel classes on COCO (\textbf{FRCN w/ R-101 C4}). * indicates that the results are averaged over 10 random seeds. $\dagger$ denotes that the results are reproduced with the official codes.}
  \label{tab:coco_main_c4}
\end{table*}

\minisection{Results on PASCAL VOC.} We evaluate our method along with previous SOTA methods and their corresponding baselines \texttt{YOLO/FRCN-ft-full}, where the whole network is finetuned until convergence during few-shot transfer.
As shown in Table~\ref{tab:main_voc}, our \texttt{PTF} baseline performs on par with the previous SOTA methods \texttt{FSCE}~\cite{sun2021fsce}, \texttt{SRR-FSD}~\cite{zhu2021semantic}, \texttt{$\text{FSOD}^{up}$}~\cite{wu2021universal}.
Incorporating with our KI initializer, \texttt{PTF+KI} surpasses all other methods across all shots and splits except for the 1-shot and 2-shot cases of Novel Set 2 as runner up.
Note that \texttt{PTF+KI} is a versatile among all shots, while \texttt{FSCE}~\cite{sun2021fsce} is preferable with more shots, \textit{e.g.}, 5-shot and 10-shot, and \texttt{SRR-FSD}~\cite{zhu2021semantic} and \texttt{$\text{FSOD}^{up}$}~\cite{wu2021universal} favor fewer shots, \textit{e.g.}, 1-shot and 2-shot.
Building upon \texttt{DeFRCN} with FRCN w/ R-101 C4, our \texttt{PTF+KI} consistently outperforms \texttt{DeFRCN}~\cite{qiao2021defrcn} especially in the extremely few-shot cases, e.g., 1-shot and 2-shot, by a large margin ($2\text{\textendash}11$ points).

Beyond the generalization capacity evaluation, we also investigate the forgetting issue in generalized FSOD by presenting thorough comparisons on the base performance under 3-shot and 10-shot cases of Novel Set 1, as shown in Table~\ref{tab:voc_base}. We can observe that \texttt{Meta} \texttt{R-CNN}~\cite{yan2019meta}, \texttt{FSDetView}~\cite{xiao2020few}, \texttt{MPSR}~\cite{wu2020multi} and \texttt{$\text{FSOD}^{up}$}~\cite{wu2021universal} severely compromise the base performance by $10\text{\textendash} 16$ points compared with \texttt{Train} \texttt{base} \texttt{only}, where the model is only trained on the base classes to serve as an upper bound. In contrast, our \texttt{PTF+KI} not only achieves SOTA results on the novel classes, but also maintains reasonable performance on the base classes (only $2\text{\textendash} 3$ points drop).
Building upon \texttt{DeFRCN}, our \texttt{PTF+KI} still manifests its superior capability in maintaining the base knowledge against \texttt{DeFRCN}~\cite{qiao2021defrcn}.

\begin{table*}[ht]
	\centering
	\footnotesize
	\setlength{\tabcolsep}{0.8em}
	\renewcommand{\arraystretch}{0.85}
	\adjustbox{width=\linewidth}{
		\begin{tabular}{l|c|ccc|ccc|ccc|c}
			\toprule
			Method & Backbone & AP & AP50 & AP75 & APs & APm & APl & APr & APc & APf & \# Iters\\\midrule
			Joint training~\cite{gupta2019lvis}  & \multirow{5}{*}{FRCN w/ R-50} & 19.8 & 33.6 & 20.4 & 17.1 & 25.9 & 33.2 &\cellcolor{Gray} 2.1 & 18.5 & \textbf{28.5}  &-\\
			TFA w/ fc~\cite{wang2020frustratingly} &  & 22.3 & \textbf{37.8} & 22.2 & 18.5 & 28.2 & 36.6 & \cellcolor{Gray}14.3 & 21.1 & 27.0 & 20k\\
			TFA w/ cos~\cite{wang2020frustratingly} &   & 22.7 & 37.2 & 23.9 & \textbf{18.8} & 27.7 & \textbf{37.1} & \cellcolor{Gray}15.4 & 20.5 & 28.4 & 20k\\
			\rowcolor{Gray} PTF (Ours)  & & 23.2& 37.4 & 24.7 & 18.1 & 28.1 & 36.0 & 16.4 & \textbf{23.7} & 25.2 & 10k\\
			\rowcolor{Gray} PTF+KI (Ours)  & & \textbf{23.4} & \textbf{37.8} & \textbf{24.8} & 18.4 & \textbf{28.8} & 36.3 & \textbf{19.6} & 22.1 & 26.5 & \textbf{2.5k}\\
			\midrule
			Joint training~\cite{gupta2019lvis}  & \multirow{5}{*}{FRCN w/ R-101} & 21.9 & 35.8 & 23.0 & 18.8 & 28.0 & 36.2 & \cellcolor{Gray}3.0 & 20.8 & \textbf{30.8} & -\\
			TFA w/ fc~\cite{wang2020frustratingly} &  & 23.9 & 39.3 & 25.3 & 19.5 & 29.5 & 38.6 & \cellcolor{Gray}16.2 & 22.3 & 28.9 & 20k \\
			TFA w/ cos~\cite{wang2020frustratingly} &  & 24.3 & 39.3 & 25.8 & \textbf{20.1} & 30.2 &  \textbf{39.5} &  \cellcolor{Gray}18.1 & 21.8 & 29.8 & 20k\\
			\rowcolor{Gray} PTF (Ours) & & 24.5 & \textbf{40.1} &  25.7 &  18.4 &  30.2 & 36.7 & 18.7 &  \textbf{24.8} & 26.4 & 10k \\
			\rowcolor{Gray} PTF+KI (Ours) & & \textbf{24.8} & 39.0 & \textbf{26.4}& 19.6 &  \textbf{31.0} &  38.3&  \textbf{21.2} & 22.7 &  28.9 & \textbf{2.5k}\\
			\bottomrule
	\end{tabular}}
	\caption{Generalized few-shot detection on LVIS. We compare our method with the joint training baseline~\cite{gupta2019lvis} and TFA~\cite{wang2020frustratingly} in terms of performance (AP) and adaptation speed (\# Iters). Our PTF+KI outperforms the other methods across different backbones, especially for the novel performance APr, and shows $8\times$ faster adaptation speed than TFA.}
	\label{tab:lvis_bench}
\end{table*}

\begin{table}[ht]
\centering
\footnotesize
\setlength{\tabcolsep}{0.7em}
\renewcommand{\arraystretch}{0.9}
\adjustbox{width=\linewidth}{
    \begin{tabular}{l|c|cc}
        \toprule
        Method  & Backbone &  Base AP & Novel AP  \\ \midrule
        FSRW~\cite{kang2019few} & & - & 5.6 \\
        MetaDet~\cite{wang2019meta} & & - & 7.1 \\
        Meta R-CNN~\cite{yan2019meta} & & - & 8.7 \\
        FSDetView$\dagger$~\cite{xiao2020few} & & 8.4 & 10.3 \\
        MPSR~\cite{wu2020multi} & & 17.1 & 9.8 \\
        TFA w/ fc~\cite{wang2020frustratingly} & & 32.0 & 9.1 \\
        TFA w/ cos~\cite{wang2020frustratingly} & & 32.4 & 9.1 \\
        FSCE$\dagger$~\cite{sun2021fsce} &  & 31.6 & 11.4 \\
        SRR-FSD~\cite{zhu2021semantic} & & - & 11.3 \\
        $\text{FSOD}^{up}$~\cite{wu2021universal} & & - & 11.0 \\
        Train base only &  & 39.6 & - \\
        \rowcolor{Gray} PTF (Ours) & & \underline{35.0} & \underline{11.7} \\
        \rowcolor{Gray} PTF+KI (Ours) & \multirow{-13}{*}{\shortstack{FRCN w/\\ R-101 FPN}} & \textbf{36.2} & \textbf{13.0} \\\midrule
        Train base only &  & 38.6 & - \\
        DeFRCN$\dagger$~\cite{qiao2021defrcn} &  & 34.0 & 16.0 \\
        \rowcolor{Gray} PTF+KI (Ours) & \multirow{-3}{*}{\shortstack{FRCN w/\\ R-101 C4}} & \textbf{35.8} & \textbf{16.9} \\
        \bottomrule
    \end{tabular}
}
\caption{Few-shot detection performance (AP@0.5:0.95) for both base and novel classes on the COCO 10-shot setting. Our approach not only exhibits the state-of-the-art generalization capacity to the novel classes, but also maintains remarkable performance on the base classes. $\dagger$ denotes that the results are reproduced with the official codes.}
\label{tab:coco_base}
\end{table}

\minisection{Results on COCO.}
In this more challenging benchmark, our approach still obtains promising results.
As shown in Table~\ref{tab:coco_main_fpn}, our simple \texttt{PTF} baseline yields comparable results with the SOTA method \texttt{FSCE}~\cite{sun2021fsce}.
Equipped with our KI initializer, \texttt{PTF+KI} outperforms \texttt{FSCE} by 1.6 and 1.0 points under the stringent overall AP metric for 10-shot and 30-shot respectively. 
Owing to the few-shot adaptation of the RPN module, our approach manifest salient advantages when it comes to the average recall (AR) metrics. For instance, \texttt{PTF+KI} achieves remarkably higher AR in terms of top 10 and top 100 predicted boxes than existing methods and almost doubles the AR of small (S) object (the last but three column in Table~\ref{tab:coco_main_fpn}) against the previous SOTA method of AR, \textit{i.e.}, \texttt{FSDetView}~\cite{xiao2020few}.
This is because the adapted RPN can provide more precise region proposals for novel objects.
Visualization evidence can be found in \ref{exp:adapt_vis}. 
In Table~\ref{tab:coco_main_c4}, we observe that our \texttt{PTF+KI} is capable to consistently surpass \texttt{DeFRCN}~\cite{qiao2021defrcn} in both 10-shot and 30-shot by 0.9 and 0.7 overall AP points respectively.
Additionally, we compare the base performance of different methods under COCO 10-shot setting in Table~\ref{tab:coco_base}.
It is noteworthy that the previous SOTA method \texttt{FSCE}~\cite{sun2021fsce} suffers from the forgetting issue more severely compared with \texttt{TFA}~\cite{wang2020frustratingly}, though \texttt{FSCE} yields better novel performance.
On the contrary, our method is able to achieve the best novel AP while not sacrificing the base performance. 
Surprisingly, our \texttt{PTF+KI} outperforms two SOTA methods \texttt{FSDetView} and \texttt{MPSR} in terms of base AP by almost $2\times$ and $4\times$ respectively, showing the remarkable capability of our approach to maintain the base knowledge.
We owe this to our appropriate adaptation strategies, dropout regularization and KI initializer.

\begin{table*}[ht]
    \centering
    \footnotesize
    \setlength{\tabcolsep}{0.7em}
    \renewcommand{\arraystretch}{0.9}
    \adjustbox{width=0.8\linewidth}{
    \begin{tabular}{l|c|c|cc|cc|c}
    \toprule
    \multirow{2}{*}{Method} & \multirow{2}{*}{\# Backbone} & \multirow{2}{*}{\# Params} & \multicolumn{2}{c|}{FLOPs} & \multicolumn{2}{c|}{Time} &  \multirow{2}{*}{\# Iters}\\
          &    &   &  FT [G] & Inf. [G] & FT [hrs] & Inf. [s] \\\midrule
    Vanilla R-CNN~\cite{Ren:2017:fasterrcnn} & & 60.64M & 252.3 & 259.3 & - & \textbf{0.068} & -\\
    Meta R-CNN*~\cite{yan2019meta} & & 60.73M & 1671.1 & 1390.0 & 0.682 & 0.564 & 1.8k \\
    FSDetView*~\cite{xiao2020few} & & 60.73M & 1170.0 & 408.5 & 0.470 & 0.185 & 2.0k \\
    MPSR~\cite{wu2020multi}  & & 82.12M & 479.9 & 290.7 & 1.053 & 0.086 & 2.0k \\
    TFA w/ fc/cos~\cite{wang2020frustratingly} & & 60.64M & 252.3 & 259.3 & 8.860 & 0.073 & 100k \\
    FSCE~\cite{sun2021fsce} & & 60.64M & 252.3 & 259.3 & 1.849 & 0.074 & 15k \\
    \rowcolor{Gray} PTF (Ours)&  & \textbf{60.32M} & \textbf{252.2} & \textbf{259.0} & 0.189 & \textbf{0.068} & 2.0k \\
    \rowcolor{Gray}PTF+KI (Ours) & \multirow{-8}{*}{\shortstack{FRCN w/\\ R-101 FPN}} & \textbf{60.32M} & \textbf{252.2} & \textbf{259.0} & \textbf{0.187} & \textbf{0.068} & \textbf{1.0k} \\\midrule
    Vanilla R-CNN~\cite{Ren:2017:fasterrcnn} & & 52.73M & 291.4 & 408.9 & - & \textbf{0.088} & - \\
    DeFRCN~\cite{qiao2021defrcn} &  & 52.74M & 291.4 & 408.9 & 1.137 & \textbf{0.088} & 6.0k \\
    \rowcolor{Gray}PTF+KI (Ours) & \multirow{-3}{*}{\shortstack{FRCN w/\\ R-101 C4}} & \textbf{52.09M} & \textbf{291.1} & \textbf{408.3} & \textbf{0.227} & \textbf{0.088} & \textbf{1.8k} \\
    \bottomrule
    \end{tabular}}

    \caption{Comparison of computational complexity, total few-shot transfer (FT) time and per image/device inference (Inf.) time. The computational complexity (FLOPs) is reported with the input size of $1200\times800\times3$.
    The FLOPs of FRCN w/ R101 C4 are calculated with the same RoI pooling resolution of $7\times 7$ as DeFRCN~\cite{qiao2021defrcn}. * indicates the method is tested on GTX 1080 Ti due to the outdated codes with cuda compatibility issues, while the others are tested on RTX 2080 Ti.}
    \label{tab:efficiency}
\end{table*}

\minisection{Results on LVIS.}
To verify the scalability capacity, we evaluate our method along with two competitors~\cite{gupta2019lvis,wang2020frustratingly} on the LVIS dataset with 1230 classes in total, including 776 base classes and 454 novel classes. As shown in table~\ref{tab:lvis_bench}, by simply adapting the base learner and RPN, our baseline framework \texttt{PTF} yields better results and approximately $2\times$ faster adaptation speed than the previous SOTA method \texttt{TFA}. Incorporating with our proposed KI initializer, \texttt{PTF+KI} surpasses all other methods by a large margin, particularly for the novel performance APr by $3\text{\textendash}5$ points, demonstrating the compelling generalization capacity of our method. 
Besides, \texttt{PTF+KI} further accelerates the adaptation speed (measured with \# Iters) by $4\times$ over \texttt{PTF}, bringing the overall speedup to $8\times$ against \texttt{TFA}. We leave the detailed definition of adaptation speed in Sec.~\ref{exp:efficiency}. All the results prove that our method not only can be readily scalable to the few-shot tasks with large-scale vocabulary, but even exhibits surprisingly fast adaptation speed.

\subsection{Efficiency Analysis}\label{exp:efficiency}
We compare the efficiency of different approaches in both few-shot transfer and inference phases. Note that the efficiency of few-shot transfer involves both computational complexity and adaptation speed, while that of inference phase only relies on computational complexity. First, we theoretically calculate the number of FLOPs involved for a single feed-forward pass to compare the computational complexity of different methods, which is invariant to different implementations.
Then we evaluate the adaptation speed by measuring the number of iterations required for convergence during few-shot transfer.
Besides, we also experimentally measure the actual time cost in both few-shot transfer and inference phases to straightforwardly reflect the overall efficiency of different methods.
\hl{Finally, since this is the first work to point out and analyze the efficiency issues in FSOD, to be comprehensive, we summarize two general overhead sources that compromise the efficiency of the popular meta-learning principle in the appendix.}

\minisection{Computational complexity.}
Practically, since the batch size and input resolution can vary across different methods, we calculate the computational complexity as the total FLOPs required for a single feed-forward model inference with a fixed input size of $1200\times 800 \times3$ to ensure fair comparison, which is commonly adopted in efficiency analysis~\cite{yang2020small,cheng2021extremely,wang2018pelee}.
Note that we calculate the FLOPs with the highest possible number of RPN proposals being processed by the subsequent RoI head, \textit{i.e.}, 512 for training and 1000 for inference by default. In practice, the actual FLOPs may be lower when the number of RPN proposals output by the RPN module does not reach the threshold.
As Table~\ref{tab:efficiency} shows, \texttt{TFA}~\cite{wang2020frustratingly}, \texttt{FSCE}~\cite{sun2021fsce}, \texttt{DeFRCN}~\cite{qiao2021defrcn}, \texttt{PTF} and \texttt{PTF+KI} bring no computational increment in both few-shot transfer and inference compared with \texttt{Vanilla R-CNN}. It is noteworthy that the FLOPs of FSCE in Table~\ref{tab:efficiency} does NOT include the overhead brought by their constrastive loss CPE, since they do not adopt the CPE loss on the COCO benchmark, which is not stated in their original paper though.
We highlight that \texttt{PTF} and \texttt{PTF+KI} share the same FLOPs, since KI initializer does not alter the network architecture and adds no computational expense.
Thanks to the concise design, both \texttt{PTF} and \texttt{PTF+KI} take up only $15.1\%$, $21.6\%$ and $52.6\%$ FLOPs of the other complicated methods \texttt{MPSR}~\cite{wu2020multi}, \texttt{FSDetView}~\cite{xiao2020few} and \texttt{Meta} \texttt{R-CNN}~\cite{yan2019meta} respectively, which introduce either extra components or inputs.
In summary, a concise architecture design can effectively avoid adding computational complexity to the base model.

\begin{figure}[t]
    \vspace{-3mm}
    \centering
    \includegraphics[width=\linewidth]{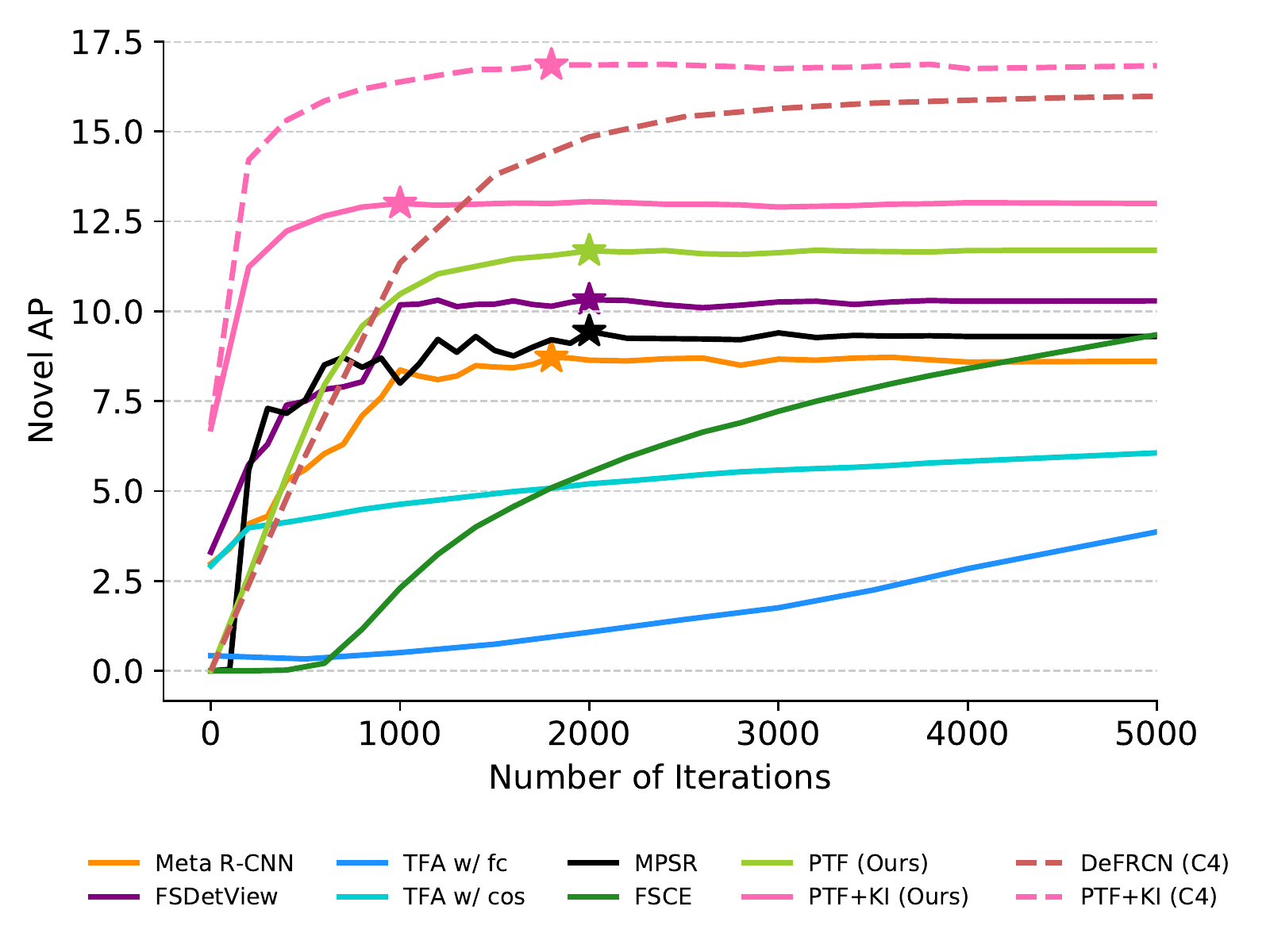}
    \caption{Comparison of few-shot adaptation speed between different methods under COCO 10-shot setting. The star indicates a model's convergent point, where the corresponding iteration is regarded as adaptation speed, based on the definition in Sec.~\ref{exp:adapt}. The dashed line indicates the method uses FRCN w/ R-101 C4 instead of FRCN w/ R-101 FPN.}
    \label{fig:adapt}
\end{figure}

\minisection{Adaptation speed.}\label{exp:adapt}
We define adaptation speed as the number of iterations (\# Iters) required for model convergence.
Concretely, we evaluate a model every 200 iterations and update the best novel AP (nAP) on the fly. If current nAP no longer surpasses the best recorded nAP for consecutive 2000 iterations, we regard that the model has converged and report the iteration that yields the best nAP as adaptation speed. The fewer iterations are required, the faster adaptation speed will be.
In Fig.~\ref{fig:adapt}, we plot the novel AP over the number of iterations under the COCO 10-shot setting to vividly compare the adaptation speed of different methods.
The exact number of iterations required for convergence can be found in the last column of Table~\ref{tab:efficiency}.
For better visualization effect, we only plot out the first 5000 iterations, as \texttt{TFA}~\cite{wang2020frustratingly} and \texttt{FSCE}~\cite{sun2021fsce} require much more (\textit{i.e.}, 100k and 15k respectively) iterations toward convergence than the other methods.
Concretely, \texttt{TFA} exhibits a drastically slow adaptation speed due to their inappropriate adaptation strategy and the length inconsistency issue in their initialization strategy for the novel weights.
Recent approach \texttt{FSCE}, though improving the adaptation speed of \texttt{TFA} from 100k to 15k by updating large proportion of the base model weights without appropriate regularizations during few-shot transfer, suffers from the catastrophic forgetting more severely.
On the contrary, our \texttt{PTF} baseline shows comparable adaptation speed with \texttt{Meta R-CNN}~\cite{yan2019meta}, \texttt{FSDetView}~\cite{xiao2020few} and \texttt{MPSR}~\cite{wu2020multi} (all around 2000 iterations).
Thanks to a better initial point, our \texttt{PTF+KI} adapts significantly fast with only 1000 iterations to reach the peak performance, achieving $1.8\text{\textendash}100\times$ adaptation speed boost compared with the other methods.
More surprisingly, only $40\%$ of the total iterations (\textit{i.e.}, 400 iterations) can achieve $94\%$ of the peak performance, which indicates that our method adapts fast especially in the early stage. This is a valuable property because it means that we can save $60\%$ training time with only $6\%$ performance drop.
Further, even without any network finetuning (\textit{i.e.}, iteration equals to zero in Fig.~\ref{fig:adapt}), \texttt{PTF+KI} can obtain a reasonable novel AP (roughly $2\times$ against two meta-learning methods \texttt{Meta R-CNN} and \texttt{FSDetView}), making our method practicable even with extremely limited computational resources, where model training is unavailable. This validates that our KI initializer can provide a good initial point and substantially reduce the number of gradient steps for convergence.
When building upon Faster R-CNN with ResNet101-C4, \texttt{PTF+KI} showcases $3.3\times$ faster adaptation speed than \texttt{DeFRCN}~\cite{qiao2021defrcn}.
On LVIS, we can similarly observe promising results in the last column of Table~\ref{tab:lvis_bench} with our \texttt{PTF+KI} gaining $8\times$ faster adaptation speed than \texttt{TFA}.

\minisection{Time cost.} As complementary to the above two metrics, time cost is a more straightforward metric that reflects the overall efficiency, which takes into account the backward propagation overhead while undesirably being sensitive to implementation variances.
From Table~\ref{tab:efficiency}, we can find that \texttt{PTF} and \texttt{PTF+KI} consume less time than the other methods during inference, which is consistent with the calculated inference computational complexity (FLOPs).
As expected, our concise network design leads to lower computational complexity and hence faster inference speed.
During few-shot transfer, \texttt{TFA}~\cite{wang2020frustratingly}, \texttt{FSCE}~\cite{sun2021fsce}, \texttt{PTF} and \texttt{PTF+KI} share almost the same feed-forward FLOPs. However, \texttt{PTF} or \texttt{PTF+KI} consumes merely 2.1\% and 10.2\% few-shot transfer time compared with \texttt{TFA} and \texttt{FSCE} respectively.
This is because \texttt{PTF} (2k) or \texttt{PTF+KI} (1k) adapts substantially faster than \texttt{TFA} (100k) and \texttt{FSCE} (15k) with much fewer iterations, owing to our improved adaptation strategies as detailed in Sec.~\ref{exp:ablation_vis}.
Finally, our \texttt{PTF+KI} takes approximately the same time for few-shot transfer compared with the \texttt{PTF} baseline, though \texttt{PTF+KI} shows $2\times$ faster adaptation speed. The underlying reason is that \texttt{PTF+KI} consumes more time on the backward propagation since it updates the backbone with a small learning rate scaling factor 0.01.

\begin{table}[!t]
\centering
\footnotesize
\setlength{\tabcolsep}{0.75em}
\adjustbox{width=0.9\linewidth}{
    \begin{tabular}{ccc|ccc}
        \toprule
        Regressor &  bAP & nAP & Classifier & bAP & nAP \\ \midrule
        Specific & \textbf{35.1} & 11.1 & Cosine & 33.2 & 11.3 \\
        Agnostic & 35.0 & \textbf{11.7} & Linear & \textbf{35.0} & \textbf{11.7}\\
        \bottomrule
    \end{tabular}
}
\caption{Comparison of different instantiations of the base learner comprising the box regressor and box classifier.}
\label{tab:base learner}
\end{table}

\begin{figure*}[t]
    \vspace{-2mm}
	\begin{center}
		\includegraphics[width=0.8\linewidth]{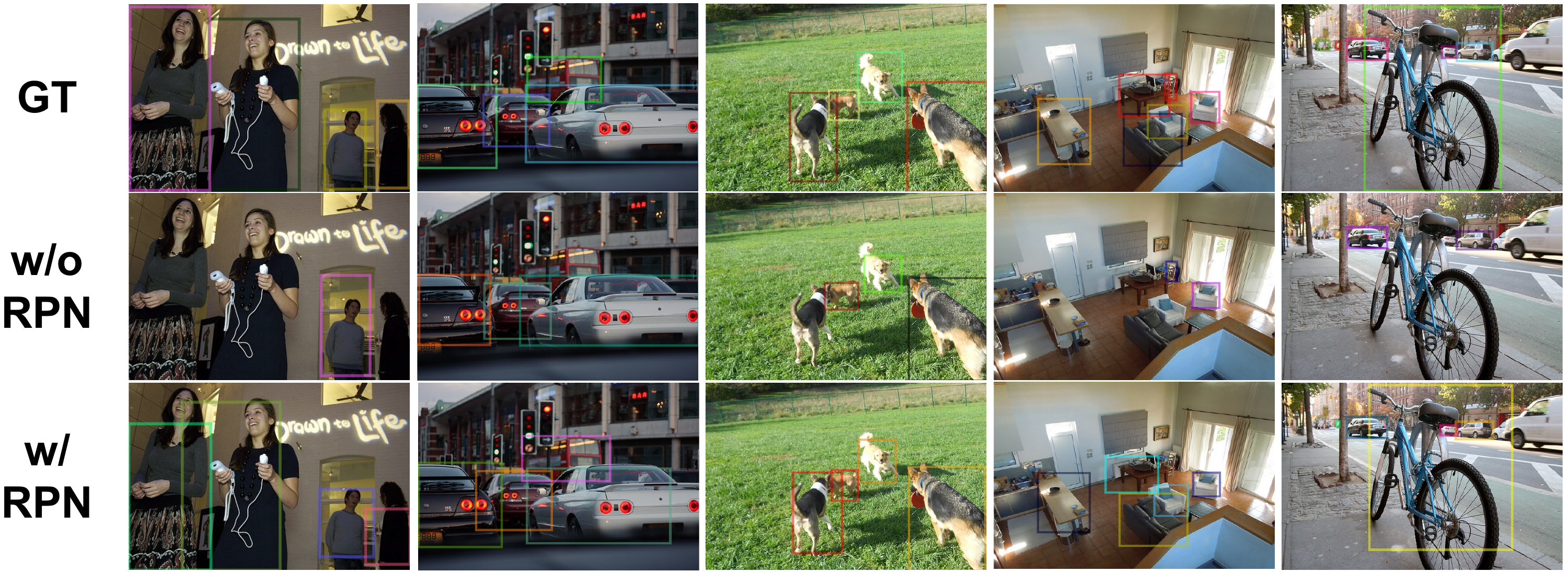}
		\caption{Visualization of the RoI proposals predicted by the RPN \textit{with} or \textit{without} adaptation during few-shot transfer. \textbf{GT} stands for the novel class ground truth boxes. \textbf{w/ RPN} and \textbf{w/o RPN} indicate whether the RPN is adapted or not. Note that we only plot the RoI proposals with the maximum IoU overlap with corresponding ground truth boxes for better visualization.}
		\label{fig:rpn}
	\end{center}
\end{figure*}

\subsection{Ablation Study and Visualization}\label{exp:ablation_vis}
Unless otherwise stated, all the ablative studies are conducted upon COCO 10-shot setting with ResNet-101 FPN~\cite{he2016deep} as backbone.

\minisection{Base learner.} Upon our \texttt{PTF} baseline, we explore the effect of different instantiations of the box regressor and box classifier respectively. As shown in Table~\ref{tab:base learner}, class agnostic box regressor outperforms class specific one in terms of novel AP. This is probably because class agnostic box regressor contains fewer parameters and can be directly re-used in few-shot transfer, which reduces the overfitting risk under few-shot scenarios.
Moreover, linear classifier yields better base and novel performance than cosine similarity based classifier, and therefore we use linear classifier throughout all experiments.

\begin{table}[t]
\centering
\footnotesize
\setlength{\tabcolsep}{0.3em}
\adjustbox{width=\linewidth}{
\begin{tabular}{cccc|ccc|ccc}
\toprule
\multirow{2}{*}{BL} & \multirow{2}{*}{RPN} & \multirow{2}{*}{RH} & \multirow{2}{*}{BB} & \multicolumn{3}{c|}{PTF} & \multicolumn{3}{c}{PTF+KI}\\
&&&& bAP & nAP & \# Iters $\downarrow$ & bAP & nAP & \# Iters $\downarrow$\\
\midrule
\cmark &  &  &  & 36.1 & 11.6 & 10k & 36.3 & 11.6 & 8k\\
\cmark & \cmark &  &  & \textbf{36.3} & 11.8 & 5k & 36.5 & 11.9 & 4k\\
\cmark &  & \cmark &  & 34.1 & 11.1 & 3k & 35.1 & 11.9 & 2k\\
\cmark & \cmark & \cmark & & \cellcolor{Gray}35.0 & \cellcolor{Gray}11.7 & \cellcolor{Gray}\textbf{2k} & 36.5 & 12.4 & \textbf{1k}\\
\cmark & \cmark & \cmark & \cmark & 34.5 & \textbf{12.3} & \textbf{2k} & \cellcolor{Gray}36.2 & \cellcolor{Gray}\textbf{13.0} & \cellcolor{Gray}\textbf{1k}\\
\bottomrule
\end{tabular}
}
\caption{Effect of different adaptation strategies during few-shot transfer. \cmark\ means this component will be finetuned. BL, RH and BB stand for base learner, RoI head and backbone respectively. The strategy highlighted in gray strikes a good balance between compelling performance and fast adaptation speed, and is therefore adopted in other experiments.}
\label{tab:adaptation}
\end{table}

\begin{table}[t]
    \centering
    \setlength{\tabcolsep}{1.3em}
    \adjustbox{width=\linewidth}{

    \begin{tabular}{cc|cccc}
    \toprule
    BL & RPN & AR & ARs & ARm & ARl \\\midrule
    \checkmark & & 15.5 & 8.8 & 13.5 & 28.8 \\
    \checkmark & \checkmark & \textbf{22.6} & \textbf{11.6} & \textbf{21.8} & \textbf{40.8} \\
    \bottomrule
    \end{tabular}}

    \caption{Effect of RPN adaptation with PTF+KI during few-shot transfer. We report the average recall (AR) of novel class objects with the top 100 RoI proposals. \checkmark means this component is adapted. BL stands for the base learner.}
    \label{tab:rpn}
\end{table}

\begin{table*}[t]
\centering
\footnotesize
\renewcommand{\arraystretch}{0.6}
\adjustbox{width=0.75\linewidth}{
\begin{tabular}{ccc|ccc|ccc}
\toprule
Frozen Block & bAP & nAP & $\alpha(\phi)$ & nAP  & bAP & Gradient Stop & nAP & bAP\\
\midrule
\cellcolor{Gray}2-nd & \cellcolor{Gray}36.2 & \cellcolor{Gray}\textbf{13.0} & 1x & 28.3 & 11.4 & \xmark & 36.0 & 12.9\\
3-rd & 36.2 & 12.9 & 0.1x & 33.8 & \textbf{13.4} & \cellcolor{Gray}\cmark & \cellcolor{Gray}\textbf{36.2} & \cellcolor{Gray}\textbf{13.0}\\
4-th & \textbf{36.6} & 12.6 & \cellcolor{Gray}0.01x & \cellcolor{Gray}\textbf{36.2} & \cellcolor{Gray}13.0  &  - & - &  -\\
\bottomrule
\end{tabular}
}
\caption{Effect of different few-shot adaptation strategies adopted exclusively in PTF+KI. Frozen Block indicates up to which residual block the backbone is frozen. $\alpha(\phi)$ is the learning rate scaling factor for the backbone.}
\label{tab:ptf_ki_adapt}
\end{table*}

\label{exp:adapt_vis}
\minisection{Adaptation strategy.} We conduct ablative experiments in Table~\ref{tab:adaptation} to analyze how different adaptation strategies affect the performance and adaptation speed during few-shot transfer. \textbf{First}, merely updating the base learner provides reasonable results but requires a long journey to converge. \textbf{Second}, finetuning the RPN together with the base learner improves the novel AP and halves the number of iterations required for convergence (\textit{i.e.}, boosting the adaptation speed by $2\times$).
We credit this gain to the improvement of RoI proposals for novel classes.
Here we provide both numbers and extensive visualization to support our claim: i) we report the average recall (AR) of novel class objects with the top 100 RoI proposals in Table~\ref{tab:rpn}. Clearly, adapting the RPN can consistently improve the novel AR across different object scales, especially the medium and large ones. ii) we visualize the RoI proposals either \textit{with} or \textit{without} RPN adaptation in Fig.~\ref{fig:rpn}. We set 0.5 as the IoU threshold to select valid RoI proposals, among which we only plot the one with the maximum IoU overlap with each novel class ground truth box for better visualization. \hl{Obviously, we can observe some novel objects missed by the unadapted RPN, \textit{e.g.}, people, bus, dog, sofa, bicycle, etc. More examples can be found in the appendix.} On the contrary, adapting the RPN can effectively alleviate this issue and generate RoI proposals for these otherwise missed novel objects.
\textbf{However}, adapting the RoI head and base learner yields poor results, especially with the \texttt{PTF} baseline. We believe this is caused by overfitting, and the randomly initialized base learner in \texttt{PTF} further enlarges the overfitting problem.
\textbf{Importantly}, fintuning the whole network except the backbone brings preferable performance and also claims the fastest adaptation speed for both \texttt{PTF} and \texttt{PTF+KI}. 
We keep this strategy for the \texttt{PTF} baseline since it strikes a reasonable balance between performance and efficiency.
\textbf{Finally}, finetuning the whole network brings further improvement on the novel performance while slightly sacrificing the base performance. 
Since \texttt{PTF+KI} requires only 1k iterations for convergence, we choose to finetune the whole network including the backbone for better novel performance with acceptable time cost increment due to the additional gradient computation for the backbone.

Additionally, we validate the adaptation strategies exclusively adopted in \texttt{PTF+KI} while not in the \texttt{PTF} baseline. As shown in Table~\ref{tab:ptf_ki_adapt}, freezing the backbone up to the second residual block yields the best novel performance. Updating the backbone with the learning rate scaling factor 0.01x achieves preferable novel performance while not sacrificing the base performance. Stopping the gradient from the RPN module slightly improves both base and novel performance.

\begin{table}[t]
\centering
\footnotesize
\setlength{\tabcolsep}{0.3em}
\adjustbox{width=\linewidth}{
\begin{tabular}{cccc|cc}
\toprule
Batch Sampl. & Agn. Reg. & Adapt. Strat. & Dropout & bAP & nAP \\
\midrule
 & & & & 32.0 & 9.1 \\
\cmark &  &  &  & 34.3 & 10.2 \\
\cmark & \cmark &  &  & 34.9 & 10.5 \\
\cmark & \cmark & \cmark & & 34.0 & 11.0 \\
\cmark & \cmark & \cmark & \cmark & \textbf{35.0} & \textbf{11.7} \\
\bottomrule
\end{tabular}
}
\caption{Ablative performance (mAP) of the PTF baseline against TFA~\cite{wang2020frustratingly} on COCO 10-shot by gradually applying the proposed strategies. Batch Sampl. refers to our image-level batch sampling scheme.
Agn. Reg. stands for the class-agnostic box regressor.
Adapt. strat. includes the scaled learning rate strategy, and the approprate adaptation strategy for different components as studied in Table~\ref{tab:adaptation}.}
\label{tab:ptf_vs_tfa}
\end{table}

\minisection{PTF baseline vs. TFA.}
We break down the improvement that our \texttt{PTF} baseline gains compared with \texttt{TFA}~\cite{wang2020frustratingly} by gradually applying our proposed strategies, as in Table~\ref{tab:ptf_vs_tfa}. Specifically, applying the \textbf{image-level batch sampling} yields promising improvement on both base and novel performance. It shows that preserving the annotation completeness of the sampled input images can avoid wrongly suppressing foreground objects as background, which is crucial to few-shot adaptation.
Moreover, replacing the class-specific box regressor with class-agnostic one slightly boosts the performance, as studied in the \textit{base learner} paragraph above.
Further, using \textbf{scaled learning rate} (large global learning rate applied) along with appropriate \textbf{adaptation strategy} for each component enhances the generalization ability and drastically boosts the adaptation speed with only 2k iterations to converge. The compromise in base performance is caused by overfitting since more components are finetuned. More details can be found in the \textit{adaptation strategy} paragraph above.
Finally, our proposed \textbf{dropout strategy} effectively alleviates the overfitting issue with base and novel performance improved by 1.0 and 0.7 respectively.
This consolidates our hypothesis that dropout strategy is capable to prevent the model from forgetting the base knowledge or overfitting on the few-shot dataset.

\begin{table}[t]
\centering
\footnotesize
\setlength{\tabcolsep}{0.8em}
\renewcommand{\arraystretch}{0.9}
\adjustbox{width=\linewidth}{
    \begin{tabular}{l|cc|cc}
        \toprule
        \multirow{2}{*}{Initializer}  & \multicolumn{2}{c|}{COCO} & \multicolumn{2}{c}{LVIS}  \\ 
        & bAP & nAP & bAP & nAP\\ \midrule
        Without & 35.0 & 11.7 & \textbf{25.5} & 18.7 \\
        Imprinted (Cos.)~\cite{qi2018low} & 35.6 & 12.9 & 18.7 & 18.6 \\
        TFA~\cite{wang2020frustratingly} & 34.8 & 12.0 & 25.4 & 18.9 \\
        KI w/o ALR (Ours) & \textbf{36.2} & 12.9 & 14.9 & \textbf{22.2} \\
        \rowcolor{Gray} KI w/ ALR (Ours) & \textbf{36.2} & \textbf{13.0} & \textbf{25.5} & 21.2 \\
        \bottomrule
    \end{tabular}
}
\caption{Designs of knowledge inheritance. Our ALR strategy (highlighted in gray) addresses the inconsistency problem and achieves SOTA results without catastrophic forgetting.}
\label{tab:KI}
\end{table}

\begin{table}[t]
\centering
\footnotesize
\renewcommand{\arraystretch}{0.8}
\adjustbox{width=0.8\linewidth}{
\begin{tabular}{c|cc|cc}
\toprule
\multirow{2}{*}{Method / Shot} & \multicolumn{2}{c|}{3} & \multicolumn{2}{c}{10} \\
& bAP & nAP & bAP & nAP \\\midrule
PTF & \textbf{77.9} & 27.0 & 76.8 & 37.3 \\
PTF+KI & \textbf{77.9} & \textbf{33.1} & \textbf{77.9} & \textbf{45.1} \\
\bottomrule
\end{tabular}
}
\caption{Effect of large semantic gap between base and novel classes on KI. Both bAP and nAP are reported with the metric AP50 under our specifically defined base/novel split on PASCAL VOC benchmark.}
\label{tab:semantic_gap}
\end{table}

\minisection{KI initializer \& ALR strategy.}
We validate the effectiveness of the proposed KI initializer and ALR strategy on COCO and LVIS benchmarks (Table~\ref{tab:KI}).
Specifically, introducing \texttt{Imprinted}~\cite{qi2018low} into our \texttt{PTF} baseline cannot yield preferable results on either COCO or LVIS.
Besides, \texttt{TFA}~\cite{wang2020frustratingly} uses the classifier weights trained on novel classes as initialization, which however does not contribute to the final performance.
Notably, when training the classifier on the few-shot novel data, the annotations for the balanced base class samples are removed.
As a result, the classifier will separate the learned novel centroids fairly apart from the base class representations (treated as background since annotations removed), though there generally exists some semantic correlations between base and novel classes.
On the contrary, our KI is capable to effectively leverage the semantic relations by extracting feature representations for novel samples using the base model.
In this way, the obtained novel representations in the embedding space will naturally be closed to those base classes that share some semantic characteristics in common.
Given the above distinction of how the novel centroids are formed between our KI and \texttt{TFA}, we hypothesis that the separation of the base and novel centroids may hinder the knowledge transfer from base classes to novel classes.
Therefore, we owe the superior initialization effect of KI to leveraging the semantic relations between base and novel classes implicitly built by representation extraction from the shared base model.

Our vanilla KI initializer (\texttt{KI} \texttt{w/o} \texttt{ALR}) performs comparable with \texttt{KI} \texttt{w/} \texttt{ALR} on COCO, while suffering from catastrophic forgetting on LVIS.
This is well-aligned with our expectation because the vanilla KI will suffer from the length inconsistency issue, which is more severe on LVIS than COCO.
Concretely, the average length of the base class centroids on COCO is 1.07 (refer to Fig.~\ref{fig:observation2} for the length data), which is close to the unit length.
In this case, ALR performs similar to the $\mathcal{L}$-2 normalization adopted by \texttt{KI} \texttt{w/o} \texttt{ALR}, except that ALR allows each individual novel class centroid to slightly deviate from the unit length as long as the average length persist in unit length, while the $\mathcal{L}$-2 normalization will strictly force each individual class centroid to be exactly in unit length.
We empirically find that both strategies work well under the above circumstance.

In contrast, the average length of the base class centroids on LVIS is 0.58, which is far away from the unit length.
In this circumstance, our ALR will rescale the novel class centroids to have the same average length of 0.58 with the base centroids, while the $\mathcal{L}$-2 normalization still insists on scaling the novel centroids to have unit length.
Consequently, $\mathcal{L}$-2 normalization will introduce the length inconsistency issue between the base centroids and the predicted novel centroids, which drastically devastates the base performance and causes catastrophic forgetting.
However, ALR can effectively tackle the inconsistency problem by adaptively scaling the novel centroids.
This explains why our \texttt{KI} \texttt{w/} \texttt{ALR} can achieve SOTA results on both COCO and LVIS benchmarks without compromising the base performance.
In summary, our ALR is a more versatile strategy compared to the $\mathcal{L}$-2 normalization.

\minisection{Large semantic gap between base and novel classes.}
In practice, there may exist a circumstance that the unseen novel classes are not similar or even irrelevant to the base classes, which we refer to as large semantic gap. In order to demonstrate the effectiveness of our KI even under this extreme scenario, we specifically design a base/novel class split with large semantic gap under the PASCAL VOC benchmark, where the novel classes are all animals with four legs, \textit{i.e.}, \{cat, dog, horse, sheep, cow\}, while the base classes are something (\textit{e.g.}, vehicles, indoor furnitures) not similar or even irrelevant to the above novel classes, \textit{i.e.}, \{aeroplane, bicycle, bird, boat, bottle, bus, car, chair, diningtable,  motorbike, person, pottedplant, sofa, train, tvmonitor\}. From Table~\ref{tab:semantic_gap}, we can see that our KI initializer consistently improves the performance on both 3-shot and 10-shot setting, especially the novel performance by 6-8 percent points. This convinces us that \textit{a trained model is able to encode some useful information of unseen objects into the embedding space, though it has never been explicitly trained to do so.} We hypothesis this is because the base classes still share some low-level features with the unseen novel classes given a large semantic gap, which may facilitate learning further discriminative features for the novel classes. For instance, like the above unseen animals, chairs and diningtables generally have four legs, though their shape and color may not be that much similar to the animals'. However, it still provides some prior knowledge when the detector inspects into the tiny details of the animals' legs to help differentiate their category. In other words, the model no longer needs to learn the basic shape of legs from scratch, but rather can focus more on the distinctive details. Overall, our KI manifests its superiority even under such an extreme scenario, where the semantic gap between base and novel classes is significant.

\section{Conclusion}
We present an efficient pretrain-transfer framework baseline without additional components, which achieves comparable results with previous state-of-the-art methods.
Upon this baseline, we build an initializer, namely knowledge inheritance (KI), to reliably initialize the novel weights for the box classifier before few-shot transfer. Within the KI initializer, we tackle the length inconsistency problem with our proposed adaptive length re-scaling (ALR) strategy.
Our approach not only achieves the state-of-the-art results across three public benchmarks, \textit{i.e.}, PASCAL VOC, COCO and LVIS, but also exhibits high efficiency with fast adaptation. 
We hope that this work can motivate a trend toward powerful yet efficient few-shot technique development.

\section*{Acknowledgement}
This research is supported by the National Research Foundation, Singapore under its AI Singapore Programme (AISG Award No: AISG-RP-2018-003), the MOE AcRF Tier-1 research grants: RG95/20, and the OPPO research grant.

\appendix

\setcounter{table}{0}
\setcounter{figure}{0}
\renewcommand{\thetable}{\Alph{table}}
\renewcommand{\thefigure}{\Alph{figure}}

\begin{figure}[t]
    \centering
	\includegraphics[width=0.9\linewidth]{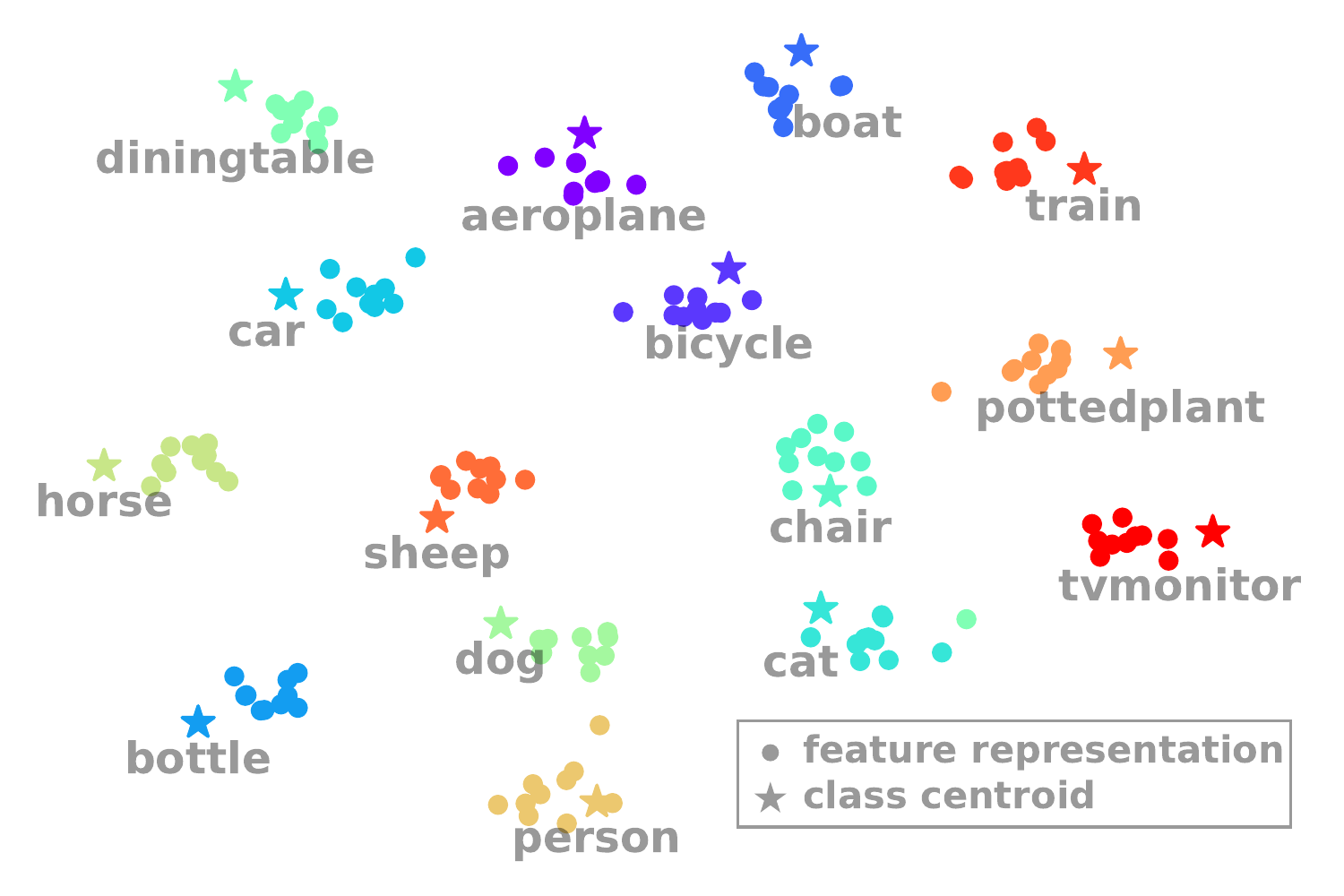}
	\caption{Visualization of the normalized base class (VOC split 1) feature representations and class centroids with t-SNE~\cite{van2008visualizing}.}
	\label{fig:observation1}
\end{figure}

\minisection{Supporting evidence for our first observation.}
To verify the first observation, we normalize the base class (VOC split 1) feature representations and class centroids to unit length and then visualize them with t-SNE~\cite{van2008visualizing} as shown in Fig.~\ref{fig:observation1}. Clearly, each class centroid and its corresponding feature representations from the same category exhibit cluster patterns after dimensionality reduction, which implies that each class centroid and its associated feature representations (both normalized) distribute closely on the unit hyper-sphere and therefore share approximately the same direction.

\begin{figure}[t]
	\begin{center}
		\includegraphics[width=\linewidth]{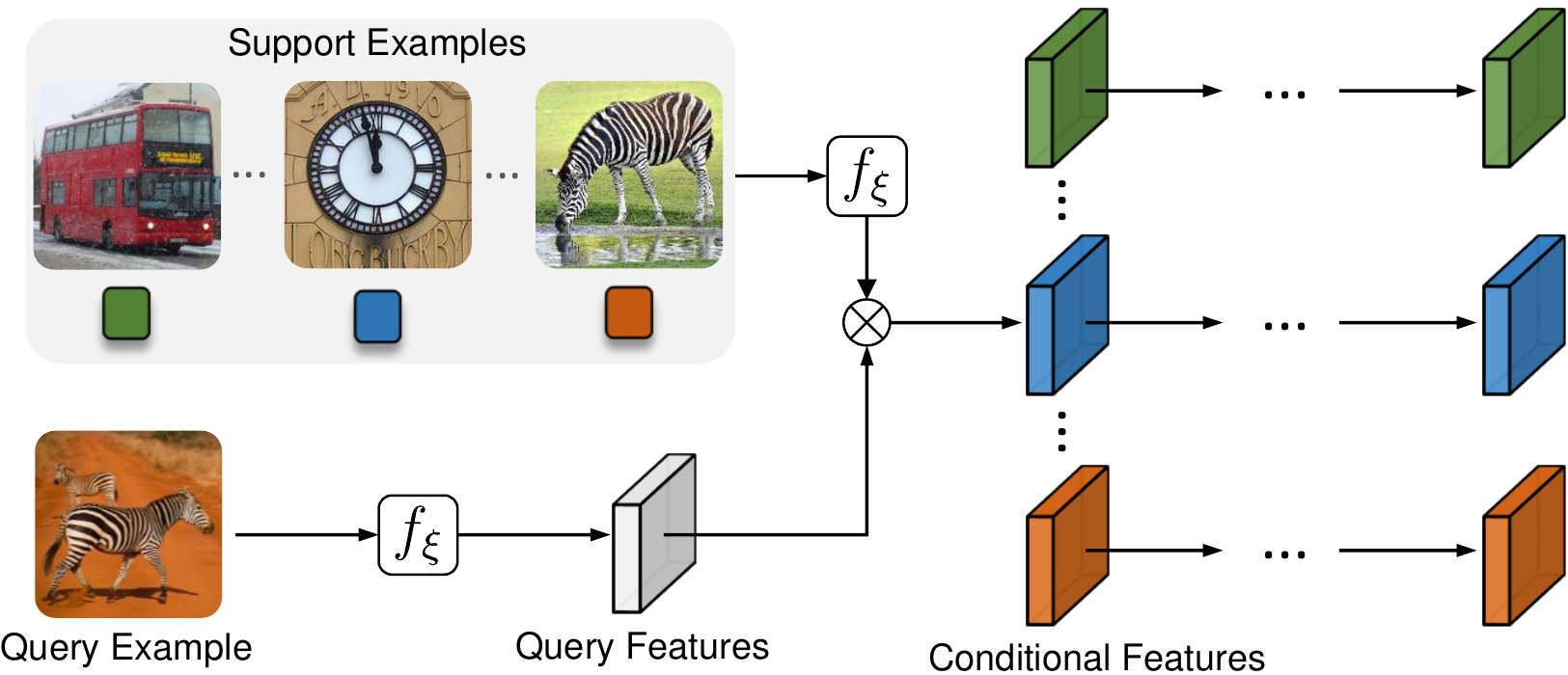}
		\caption{Illustration of the conditional divergence issue in the meta-learning principle.}
		\label{fig:diverge}
	\end{center}
\end{figure}

\minisection{Overhead sources in meta-learning.} We elaborately explain the two overhead sources of meta-learning principle, \textit{i.e.}, extra support branch and conditional divergence. Straightforwardly, an extra support branch incurs additional overheads during support feature extraction, which is proportional to the number of ways $N$ and shots $K$. On the other hand, conditional divergence
, as shown in Fig.~\ref{fig:diverge}, 
refers to the phenomenon that a union query feature representation 
(gray) 
will diverge into multiple conditional feature representations
(colored) 
once it is conditioned upon
(denoted as $\otimes$) 
different support classes.
The overheads caused by conditional divergence is proportional to the number of ways $N$ and meanwhile is also relevant to the merging stage.
Early-stage merging, \textit{e.g.}, \texttt{Meta} \texttt{R-CNN}, usually brings more overheads than late-stage merging, \textit{e.g.}, \texttt{FSDetView}, due to the larger scale feature maps and prolonged parallel computational paths after divergence.

\begin{figure*}[t]
	\begin{center}
		\includegraphics[width=0.95\linewidth]{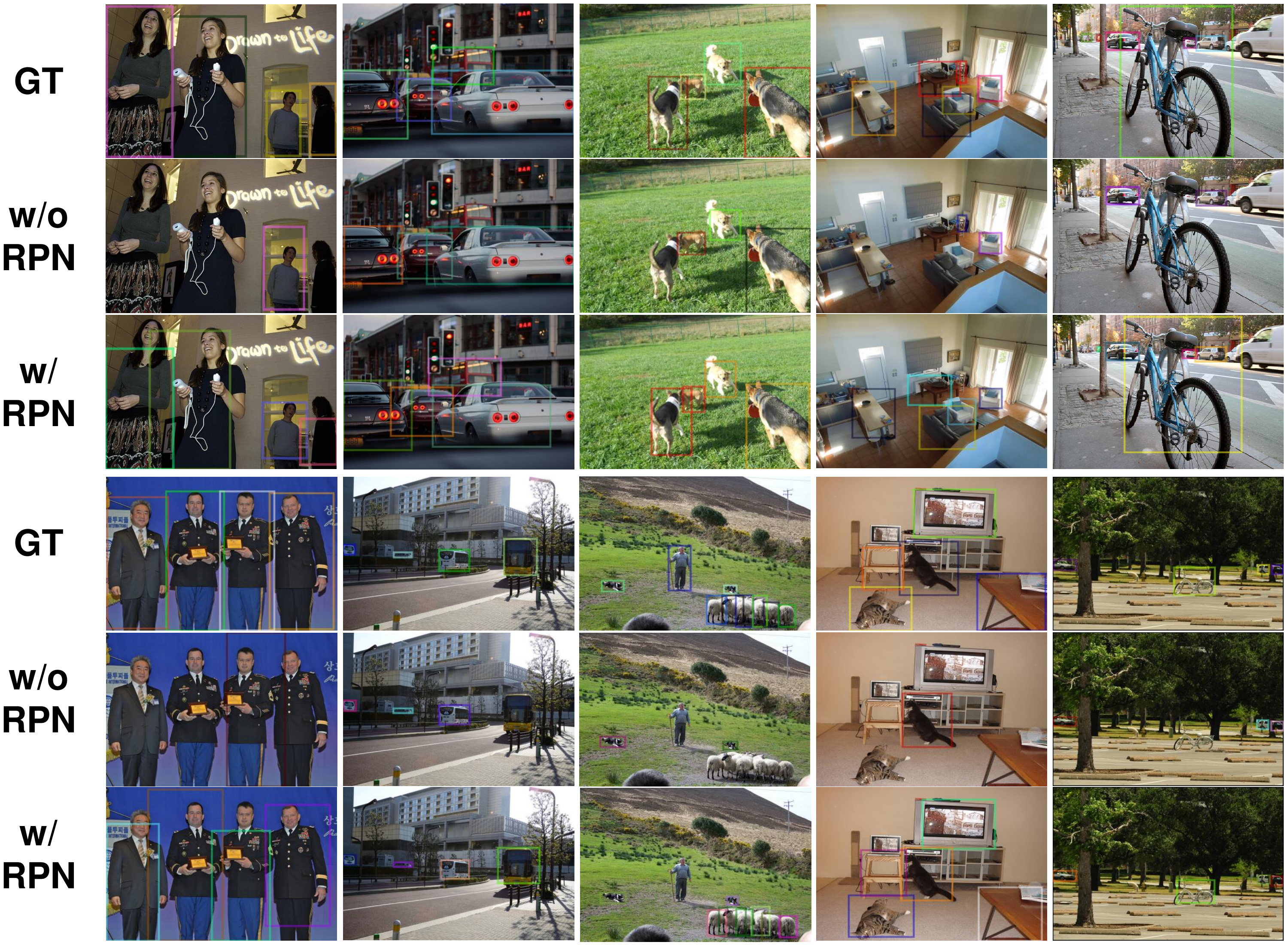}
		\caption{Visualization of the RoI proposals predicted by the RPN \textit{with} or \textit{without} adaptation during few-shot transfer. \textbf{GT} stands for the novel class ground truth boxes. \textbf{w/ RPN} and \textbf{w/o RPN} indicate whether the RPN is adapted or not. Note that we only plot the RoI proposals with the maximum IoU overlap with corresponding ground truth boxes for better visualization.}
		\label{fig:rpn}
	\end{center}
\end{figure*}

\begin{table}[t]
    \centering
    \setlength{\tabcolsep}{0.45em}
    \adjustbox{width=\linewidth}{
    
    \begin{tabular}{l|ccc|ccc}
    \toprule
    \multirow{2}{*}{Method} & \multicolumn{3}{c|}{Few-Shot Transfer} & \multicolumn{3}{c}{Inference}\\
    & FLOPs & SB & CD &  FLOPs & SB & CD \\\midrule
    Vanilla R-CNN~\cite{Ren:2017:fasterrcnn} & 252.3 & - & - & 259.3 & - & -\\
    Meta R-CNN~\cite{yan2019meta} & 1671.1 & 840.0 & 578.8 & 1390.0 & - & 1130.7 \\
    FSDetView~\cite{xiao2020few} & 1170.0 & 841.1 & 76.6 & 408.5 & - & 149.2 \\
    \bottomrule
    \end{tabular}}

    \caption{The overhead source analysis of the meta-learning based methods. We calculate the computational increment caused by the support branch (SB) and conditional divergence (CD) respectively, compared with vanilla R-CNN. The unit of the results is GFLOPs.}
    \label{tab:source}
\end{table}

Further, we analyze the overhead composition of the meta-learning based methods~\cite{yan2019meta,xiao2020few} in both few-shot transfer and inference phases (Table~\ref{tab:source}).
During few-shot transfer, the overheads brought by the support branch of both methods~\cite{yan2019meta,xiao2020few} account for over half of the totals, outlining a severe efficiency issue induced by the extra support branch in meta-learning principle. During inference, the conditional divergence issue becomes more significant since the number of instance-level RoI proposals doubles compared with few-shot transfer, which further enlarges the burden brought by conditional divergence and substantially compromises the inference speed of the few-shot detector. Finally, as expected, \texttt{Meta} \texttt{R-CNN} suffers much more ($\sim7\times$) overheads caused by conditional divergence than \texttt{FSDetView}, since \texttt{Meta} \texttt{R-CNN} features an early-stage merging.

\minisection{Visualization of RPN proposals.} We visualize the RoI proposals either \textit{with} or \textit{without} RPN adaptation in Fig.~\ref{fig:rpn}. We set 0.5 as the IoU threshold to select valid RoI proposals, among which we only plot the one with the maximum IoU overlap with each novel class ground truth box for better visualization. Obviously, we can observe some novel objects missed by the unadapted RPN, \textit{e.g.}, people, bus, sheep, dog, sofa, bicycle, cat, etc. On the contrary, adapting the RPN can effectively alleviate this issue and generate RoI proposals for these otherwise missed novel objects.

\ifCLASSOPTIONcaptionsoff
  \newpage
\fi

\bibliographystyle{IEEEtran}
\bibliography{IEEEabrv,mybibfile}

\begin{IEEEbiography}[{\includegraphics[width=1in,height=1.25in,clip,keepaspectratio]{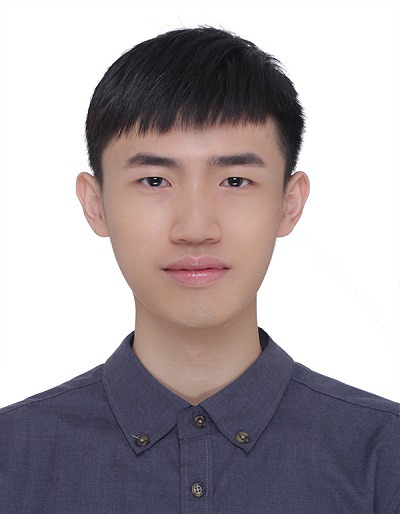}}]{Ze Yang}
is a PhD candidate with the School of Computer Science and Engineering, Nanyang Technological University (NTU). He received his B.Eng. degree from Tianjin University, China in 2019. His current research interests include few-shot learning and continual learning in computer vision.
\end{IEEEbiography}
\begin{IEEEbiography}[{\includegraphics[width=1in,height=1.25in,clip]{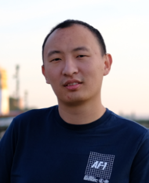}}]{Chi Zhang} is a PhD candidate with the School of Computer Science and Engineering, Nanyang Technological University, Singapore. He received the B.S. degree from China University of Mining and Technology in 2017.
His research interests are in computer vision and machine learning.
\end{IEEEbiography}
\begin{IEEEbiography}[{\includegraphics[width=1in,height=1.25in,clip,keepaspectratio]{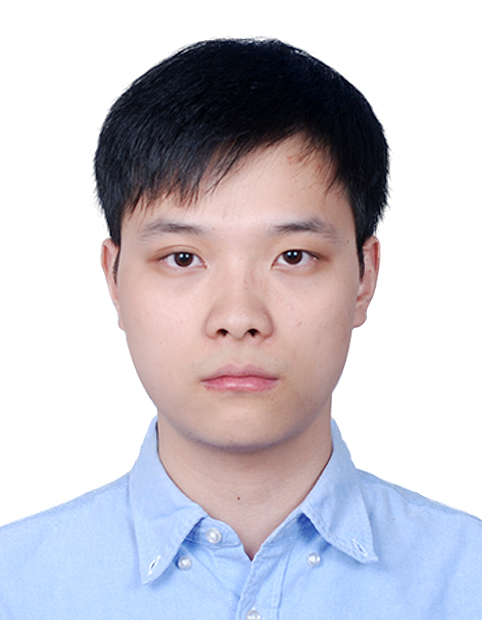}}]{Ruibo Li}
is a PhD candidate with the School of Computer Science and Engineering, Nanyang Technological University (NTU). He received his B.Eng. degree and M.S. degree from Huazhong University of Science and Technology, Wuhan, China in 2016 and 2019 respectively. His research interests are in computer vision and machine learning.
\end{IEEEbiography}
\begin{IEEEbiography}[{\includegraphics[width=1in,height=1.25in,clip,keepaspectratio]{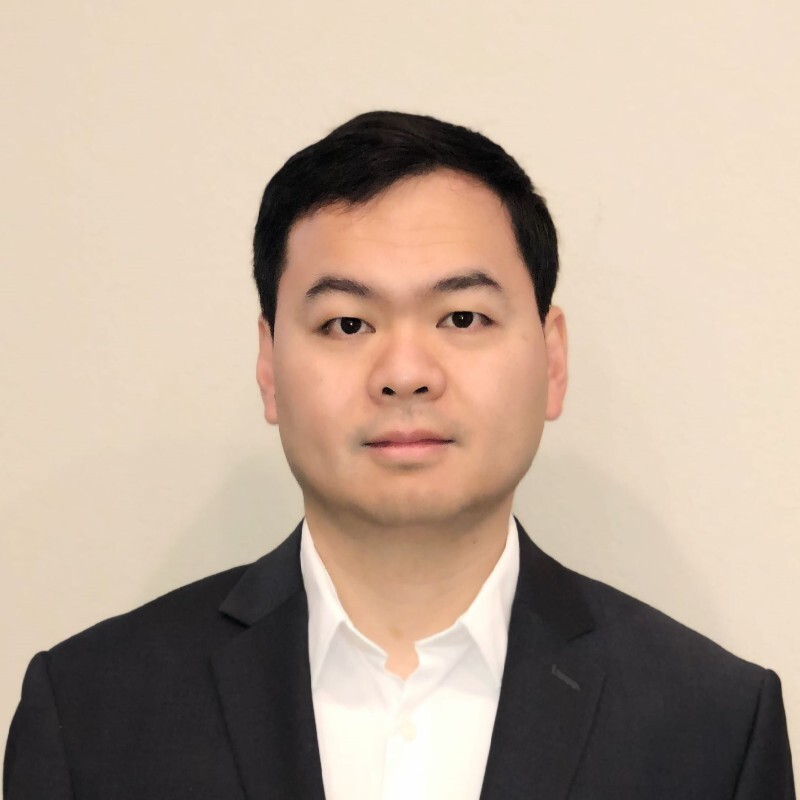}}]{Yi Xu}
is the Director of XR Technology at OPPO US Research Center, InnoPeak Technology Inc. His research interest lies in 3D computer graphics and computer vision, with a focus on Extended Reality. Before joining InnoPeak, Dr. Xu worked at various industrial labs such as GE Research and JD.COM Silicon Valley Labs. Dr. Xu got his Ph.D. degree from Purdue University in 2010.
\end{IEEEbiography}
\begin{IEEEbiography}[{\includegraphics[width=1in,height=1.25in,clip]{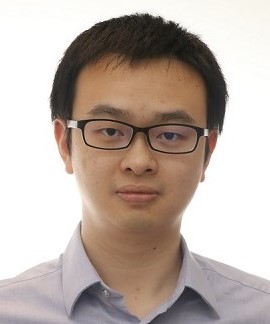}}]{Guosheng Lin} is an Assistant Professor at School of Computer Science and Engineering, Nanyang Technological University, Singapore. His research interests are in computer vision and machine learning.
\end{IEEEbiography}

\end{document}